%% file: acl_latex.tex
\pdfoutput=1

\documentclass[11pt]{article}

\usepackage[final]{acl}

\usepackage{times}
\usepackage{latexsym}

\usepackage[T1]{fontenc}

\usepackage[utf8]{inputenc}

\usepackage{microtype}

\usepackage{inconsolata}

\usepackage{graphicx}

\usepackage{todonotes}
\usepackage[whole]{bxcjkjatype}
\usepackage{booktabs}
\usepackage{multicol}
\usepackage{multirow}
\usepackage{tcolorbox}
\usepackage{amsmath}
\usepackage{latexsym}
\usepackage{amssymb}
\usepackage{xcolor}
\usepackage{subcaption}

\definecolor{customblue}{HTML}{1400AD}
\newcommand{\blue}[1]{\textcolor{customblue}{#1}}

\definecolor{customred}{HTML}{AD0000}

\definecolor{redbox}{HTML}{FFD6D6}
\newcommand{\redbox}[1]{\colorbox{redbox}{#1}}

\definecolor{bluebox}{HTML}{D6E2FF}
\newcommand{\bluebox}[1]{\colorbox{bluebox}{#1}}

\title{Understanding the Impact of Confidence in Retrieval Augmented Generation: A Case Study in the Medical Domain}
\author{%
  Shintaro Ozaki${}^{\clubsuit}$ \hspace{3pt} Yuta Kato${}^{\spadesuit}$ \hspace{3pt} Siyuan Feng${}^{\spadesuit}$ \hspace{3pt} Masayo Tomita${}^{\spadesuit}$ \hspace{3pt} Kazuki Hayashi${}^{\clubsuit}$ \hspace{3pt} \\[2pt]
   \textbf{Wataru Hashimoto}${}^{\clubsuit}$
  \textbf{Ryoma Obara${}^{\diamondsuit}$} \hspace{3pt}
  \textbf{Masafumi Oyamada${}^{\diamondsuit}$} \hspace{3pt} \\ [2pt]
  \textbf{Katsuhiko Hayashi${}^{\spadesuit}$} \hspace{3pt}
  \textbf{Hidetaka Kamigaito${}^{\clubsuit}$} \hspace{3pt} 
  \textbf{Taro Watanabe${}^{\clubsuit}$} \\ [3pt]
  ${}^{\clubsuit}$Nara Institute of Science and Technology (NAIST) \\ [2pt]
  ${}^{\spadesuit}$The University of Tokyo \hspace{3pt} 
  ${}^{\diamondsuit}$NEC Corporation \\ [2pt]
\texttt{\{ozaki.shintaro.ou6, kamigaito.h, taro.watanabe\}@naist.ac.jp}
   \\
}

\begin{document}
\maketitle
\begin{abstract}
Retrieval Augmented Generation (RAG) complements the knowledge of Large Language Models (LLMs) by leveraging external information to enhance response accuracy for queries. 
This approach is widely applied in several fields by taking its advantage of injecting the most up-to-date information, and researchers are focusing on understanding and improving this aspect to unlock the full potential of RAG in such high-stakes applications.
However, despite the potential of RAG to address these needs, the mechanisms behind the confidence levels of its outputs remain underexplored.
Our study focuses on the impact of RAG, specifically examining whether RAG improves the confidence of LLM outputs in the medical domain.
We conduct this analysis across various configurations and models.
We evaluate confidence by treating the model's predicted probability as its output and calculating several evaluation metrics which include calibration error method, entropy, the best probability, and accuracy.
Experimental results across multiple datasets confirmed that certain models possess the capability to judge for themselves whether an inserted document relates to the correct answer. 
These results suggest that evaluating models based on their output probabilities determine whether they function as generators in the RAG framework.
Our approach allows us to evaluate whether the models handle retrieved documents.\footnote{The code is available at \url{https://github.com/naist-nlp/CC_RAG}.}
\end{abstract}

\section{Introduction}
Retrieval Augmented Generation (RAG)~\cite{lewis2020retrieval} serves as a method to not only mitigate hallucinations but also supplement the knowledge of Large Language Models (LLMs)~\cite{achiam2023gpt, dubey2024llama, aizawa2024llm}. 
By leveraging external information, RAG enhances response accuracy and alignment with queries, making it widely applicable in industries. 
Notable domains include finance~\cite{yepes2024financial, setty2024improving} and healthcare~\cite{xiong-etal-2024-benchmarking}, where the reliability of information is critical.
This study focuses on the medical domain, which has relatively more text data than other fields and involves complex factors directly related to the human body.~\cite{sohn2024rationale, jeong2024improving}

\begin{figure}[t]
    \centering
    \includegraphics[width=\linewidth]{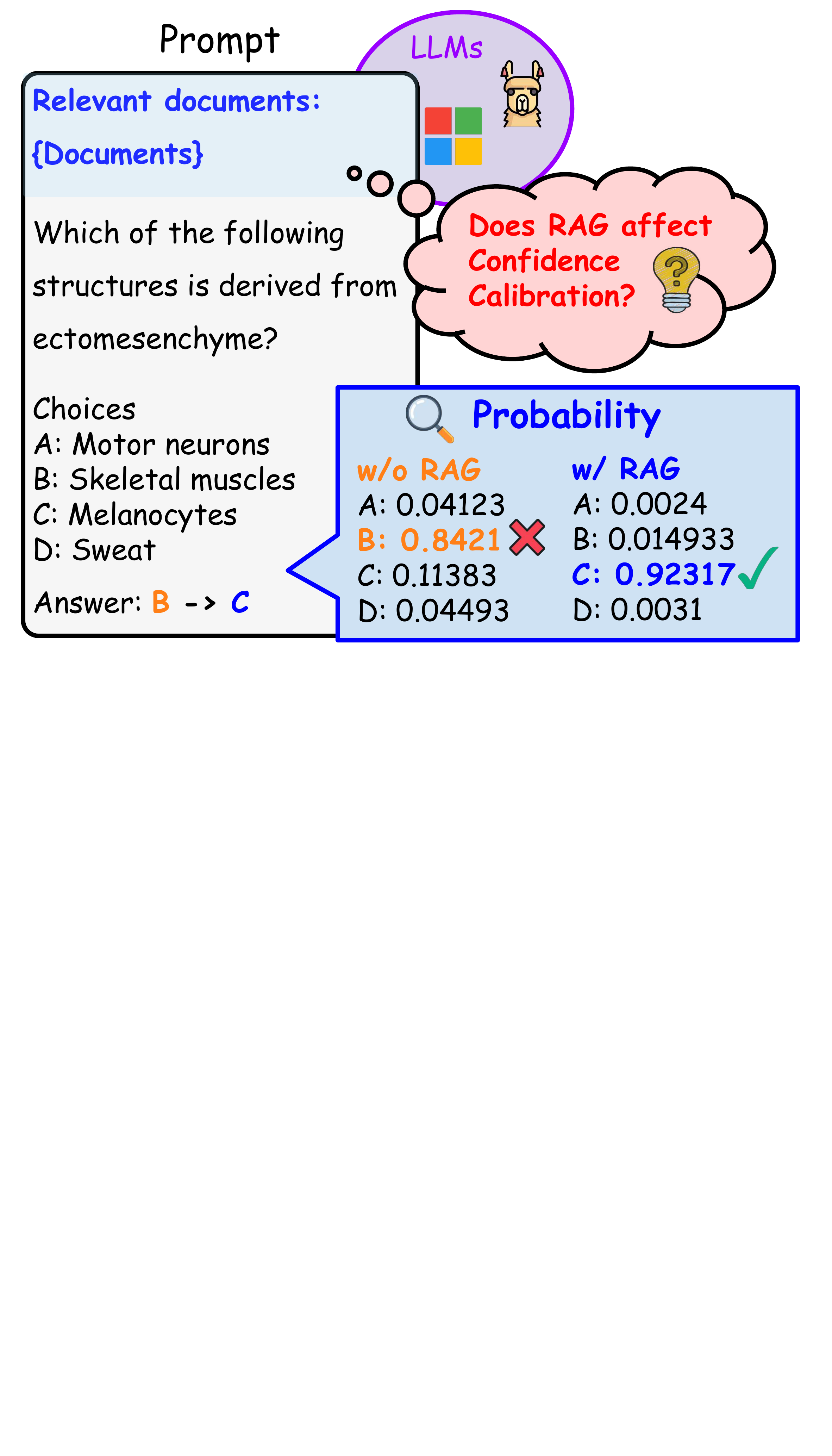}
    \caption{
    The focus of our research is to analyze whether RAG improves the confidence of the model response.
    }
    \label{fig:top_rq}
\end{figure}

While researchers explore performance improvements for LLMs using RAG, as illustrated in Figure~\ref{fig:top_rq}, analyses focusing on prediction confidence remain limited. 
Although RAG enhances answer accuracy, it may lead to overconfidence, where models exhibit unwarranted self-assurance~\cite{chen-etal-2024-controlling}.
We hypothesize that retrieving documents to support the correct answer through RAG improve the model's confidence, potentially leading to errors in confidence calibration.
Based on this, we pose a research question:
Do LLMs improve the confidence for outputs with RAG?

To address this question, we conduct a systematic analysis of multiple tasks and models in the medical domain, exploring diverse scenarios using PubMedQA~\cite{jin-etal-2019-pubmedqa} and MedMCQA~\cite{pmlr-v174-pal22a}.
In particular, we create pseudo-RAG to manipulate document content -- such as adding irrelevant documents deliberately or including only those directly related to the answer -- to simulate the range of situations RAG encounter.

Our result shows that inserting documents deliberately containing answer-supporting information improve confidence in many models, aligning with expectations, although some models exhibited behavior contrary to this prediction. 
Additionally, inserting documents unrelated to the correct answer rarely improve the confidence, suggesting that LLMs can discriminate whether an inserted document relates to the answer.
These results indicate that evaluating models based on output probabilities can lead to reveal the suitable generator model.

\begin{table}[t]
    \small
    \centering
    \begin{tabular}{ccc}
    \toprule
    \textbf{Dataset} & \textbf{Size} & \textbf{Option} \\
    \midrule
    PubMedQA & 1,000 & 3 \\
    MedMCQA (Extract) & 2,206 & 4 \\
    \bottomrule
    \end{tabular}
    \caption{
    The dataset used in our study. We select datasets that not only contain QA pairs but also include explanatory passages that justify the answers.
    }
    \label{tab:the_number_of_each_qa_datasets}
\end{table}

\section{Related Work}
\subsection{Confidence of LLMs}
Research on confidence has been prevalent since before the era of LLMs~\cite{jiang-etal-2021-know} and continues to be extensively explored~\cite{geng-etal-2024-survey}. 
\citet{becker2024cyclesthoughtmeasuringllm} proposed a framework that measures confidence by leveraging explanation-generating text produced by LLMs.
\citet{zhao2021calibrate} identified the issue that few-shot prompting significantly impacts model confidence and alters its inherent performance, and they proposed methods to address this problem. 
Confidence estimation is used as a technique to suppress 
hallucinations, where models generate false information~\cite{zhang-etal-2023-sac3}. 
\citet{cole-etal-2023-selectively} demonstrated that by utilizing model confidence, it is possible to suppress outputs for ambiguous questions. 
Our study contributes to this body of research by specifically analyzing how RAG influences confidence calibration in LLM outputs. 
Unlike prior works that primarily optimize retrieval mechanisms, we directly investigate confidence calibration dynamics.

\subsection{Boosting RAG with Confidence}
Recent advances in RAG have leveraged model confidence (e.g., output probability) to optimize retrieval and generation processes. 
For instance, \citet{jiang-etal-2023-active} introduced FLARE, which dynamically decides whether to retrieve additional information based on token-level confidence during generation, ensuring efficient retrieval by minimizing unnecessary searches.
Similarly, query rewriting techniques using reinforcement learning~\cite{ma-etal-2023-query} and strategies such as Recitation-Augmented Generation~\cite{sun2023recitationaugmented}, which searches for text resembling hypothetical answers, have shown promise in enhancing retrieval accuracy.
Moreover, recent studies like Self-RAG~\cite{asai2023self} integrate retrieval into the generation process itself.
In many of these approaches, confidence plays a crucial role either in deciding when to retrieve or in re-ranking retrieved documents based on their relevance.
However, these studies focus on improving RAG performance without analyzing how confidence itself is influenced by the RAG. 
Specifically, while confidence thresholds and re-ranking mechanisms are employed to optimize retrieval and generation, the underlying dynamics of confidence calibration within the RAG pipeline remain underexplored. 
Our study analyzes confidence calibration with and without RAG to address this gap, verify the implicit assumptions of prior works, and contribute to a deeper understanding of confidence-based mechanisms in RAG.

\section{Methods}
\label{methods}
Our study analyzes whether the confidence improves through RAG by calculating the model's confidence or entropy from the predicted probability by the model.
Each input is formatted by concatenating a system prompt, a question prompt, and its answer options (e.g., a four-choice question), following the design of Medical Information Retrieval Augmented Generation Evaluation (MIRAGE)~\cite{xiong-etal-2024-benchmarking}.
We also analyze the optimal position for inserting documents retrieved via pseudo-RAG, i.e., inserting a document directly relevant to the answer or irrelevant deliberately into the model input prompt.
Specifically, we evaluate three insertion patterns: before the question  (Pre-Question, denoted as Pre-Q), between the question and the answer choices (After-Question, denoted as Aft-Q), and after the answer choices (After-Choice, denoted as Aft-C).
This setup allows us to examine the Lost-in-the-Middle phenomenon~\cite{liu-etal-2024-lost}, where models tend to overlook intermediate content when processing long-context inputs.
Moreover, in order to focus on the impact of retrieved document positions, we use documents that contain the correct answer to the question. 
We validate our research question under three scenarios:
(1): inserting only the explanation related to the answer (denoted as Ans1). 
(2): combining the correct explanation with two irrelevant documents (denoted as Ans1-Oth2).
(3): inserting three irrelevant documents (denoted as Oth3). 
The irrelevant documents are selected from unrelated questions, ensuring that they do not contain the correct answer or semantically similar content.

Directly generating the choice answer by the model complicates evaluation, because differences in reported metrics arise even under identical conditions across studies~\cite{xiong-etal-2024-benchmarking, chen2023meditron70bscalingmedicalpretraining, wu2023pmcllamabuildingopensourcelanguage}.
In some studies, researchers select the final candidate using regular expressions, while in others, they treat the output of a specific word (such as Yes or No) as the correct answer. 
Thus, evaluation methods are not uniquely defined if the sentence generated.
In our study, we predict the most plausible option from the given choices as follows:

\[
\begin{aligned}
v_i &= \log P(x_i \mid \texttt{prompt}) \\
P(x_i) &= \frac{\exp(v_i)}{\sum_{j=1}^{J} \exp(v_j)}
\end{aligned}
\]

where \( v_i \) represents the log probability corresponding to each choice \( x_i \) and the \(\texttt{prompt}\) refers to the provided question or context.  
\( P(x_i) \) denotes the probability that the choice \( x_i \) is the correct answer, normalized by dividing the exponential of \( v_i \) by the sum of exponentials of all \( v_j \) values, while \( J \) is the number of options, which is 3 or 4.

\section{Experimental Setup}
\label{experimental_setup}
\subsection{Datasets}
We focus on the application of RAG in the medical domain.  
For the dataset, we select PubMedQA~\cite{jin-etal-2019-pubmedqa} and MedMCQA~\cite{pmlr-v174-pal22a}, both of which include multiple-choices QA data along with explanatory passages that justify the answers.  
These datasets follow the experimental setup of MIRAGE~\cite{xiong-etal-2024-benchmarking}, as shown in Table~\ref{tab:the_number_of_each_qa_datasets}.
For MedMCQA, we extract only the questions that include supporting evidence for the answer, resulting in a total of 2,206 instances.

\subsection{Inference Models}  
Following prior research~\cite{xiong-etal-2024-benchmarking}, we select the following models for evaluation: Phi-3.5 (3.8B)~\cite{abdin2024phi}, PMC-Llama (13B)~\cite{wu2023pmcllamabuildingopensourcelanguage}, Llama2 (70B)~\cite{touvron2023llama2}, LLaMA3.1 (8B / 70B)~\cite{dubey2024llama}, and Meditron (70B)~\cite{chen2023meditron70bscalingmedicalpretraining}.  
To ensure fair evaluation across models with different architectures and parameter sizes, we also include Gemma2 (2B)~\cite{team2024gemma} and Qwen2.5 (14B / 70B)~\cite{yang2024qwen2}, bringing the total to nine models.  
PMC-Llama is fine-tuned on medical domain documents based on Llama~\cite{touvron2023llama}, while Meditron undergoes continual pretraining on Llama2~\cite{touvron2023llama2}.  
For 70B models, we apply 4-bit quantization, and for PMC-Llama, we use half-precision quantization to compute probabilities.  
Detailed model configurations are provided in Appendix~\ref{app: detailed_model_setttings}.

\begin{figure}
    \centering
    \includegraphics[width=\linewidth]{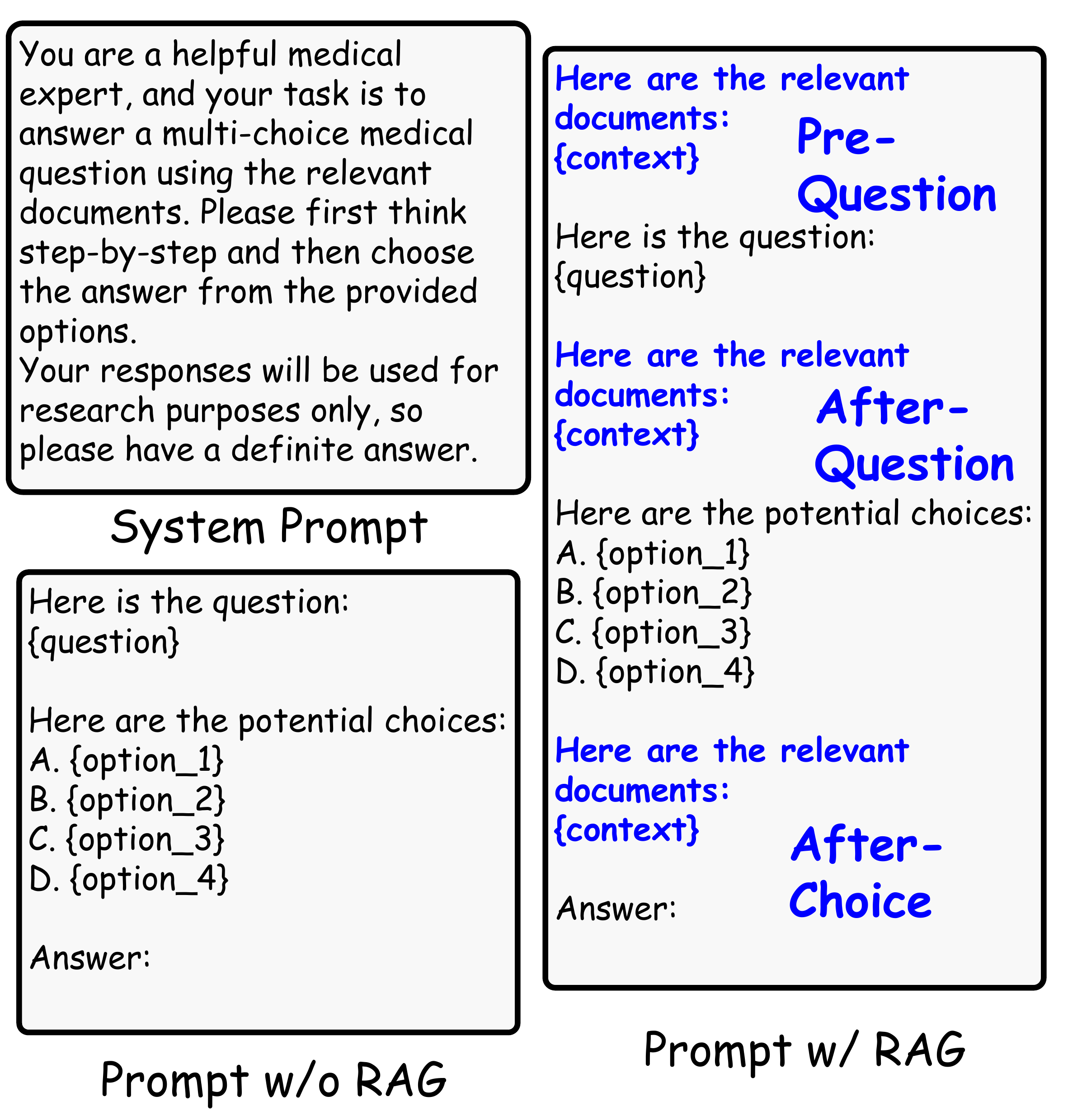}
    \caption{
    Prompts used in our research.
    Each prompt begins with a concatenated of the system prompt.
    Following MIRAGE~\cite{xiong-etal-2024-benchmarking}, we design the templates to enable the calculation of probabilities.
    }
    \label{fig:prompt_pattern}
\end{figure}

\subsection{Templates}  
Our study modifies the approach based on the MIRAGE paper~\cite{xiong-etal-2024-benchmarking} by excluding Chain of Thought (CoT)~\cite{wei2022chain}, allowing direct probability computation. 
(In other words, when using CoT, the model must generate responses, which, as discussed in Section~\ref{methods}, prevents a valid evaluation.)
Figure~\ref{fig:prompt_pattern} presents the prompts used in our study.
Each prompt incorporates system prompts from prior research~\cite{xiong-etal-2024-benchmarking} at the beginning of the input prompt. 
To investigate whether the Lost-in-the-Middle phenomenon~\cite{liu-etal-2024-lost}, also occurs in RAG, our study inserts retrieved documents at three positions: before the question (Pre-Q), after the question (Aft-Q), and after the answer choices (Aft-C).

\subsection{Evaluation Metrics}
\label{evaluation-metrics}
We evaluate if RAG boosts LLM confidence using entropy, best probability, accuracy, and Adaptive Calibration Error. 
In our multiple-choice QA task, each question has one correct answer, and output probabilities classify responses as correct or not.

\paragraph{Entropy.} \par  
We examine how entropy changes for candidate answer choices under the influence of RAG, calculating an entropy.
Ideally, inserting an answer-containing document should decrease entropy (indicating a more confident selection of the correct choice), while inserting entirely unrelated documents should improve entropy.
The entropy is computed as:

\[
H(P) = -\sum_{i=1}^{J} P(x_i) \log P(x_i)
\]

\[
\begin{aligned}
P(x_i) &= \frac{\exp(v_i)}{\sum_{j=1}^{J} \exp(v_j)}
\end{aligned}
\]

Here, $x_i$ represents a candidate answer among $J$ total options, and $v_i$ denotes the logit score (i.e., the unnormalized log-probability) assigned to $x_i$. 
The softmax function transforms these logits into a probability distribution $P(x_i)$, from which the entropy $H(P)$ is calculated. Lower entropy indicates higher model confidence in a particular choice, while higher entropy implies uncertainty.

\paragraph{Best Probability.} \par
We define ``Best Probability'' as the highest output probability among the candidate choices given to the model. 
In our study, we evaluate this metric as confidence. 
A high output probability shows strong confidence for correct answers, while a low output probability is preferred for incorrect answers (we want irrelevant documents to lower the model's confidence).

The notation of best probability is as follows:

\[
x^{*} = \arg\max_{x_i \in \mathcal{X}} \left( \log P(x_i \mid \texttt{prompt}) \right)
\]

\[
\begin{aligned}
P(x_i) &= \frac{\exp(v_i)}{\sum_{j=1}^{J} \exp(v_j)}
\end{aligned}
\]

Here, $\mathcal{X}$ is the set of all candidate answer choices, and $x^*$ denotes the choice with the highest log-probability.
Each $v_i$ represents the model's logit for the candidate $x_i$, and the softmax function converts these logits into a probability distribution over all choices.
The selected $x^*$ corresponds to the most confident prediction the model makes under the given prompt.
This Best Probability reflects how strongly the model favors its top prediction, and it serves as an interpretable confidence score in our evaluations.

\paragraph{Adaptive Calibration Error (ACE).} \par
Adaptive Calibration Error (ACE)~\cite{Nixon_2019_CVPR_Workshops} is a metric proposed to address the shortcomings of Expected Calibration Error (ECE)~\cite{naeini2015obtaining, guo2017calibration}, specifically aiming to reduce the risk of bins with a small number of samples.
~\citet{proskurina-etal-2024-quantization} and ~\citet{ulmer-etal-2022-exploring} have pointed out that ACE is a more suitable calibration error metric for multi-class classification problems. 
Based on these findings, we adopt ACE in our evaluation.
Table~\ref{tab:exp-settings} provides a complete list of all combinations and Appendix~\ref{app:deetails-of-evaluation-metrics} the details of evaluation metrics.

\begin{table}[t]
    \small
    \centering
    \resizebox{\linewidth}{!}{
    \begin{tabular}{cc}
    \toprule
    \textbf{Settings} & \textbf{Options} \\
    \midrule
    QA & PubMedQA, MedMCQA \\ [2pt]
    \multirow{2}{*}{Model} & Gemma2, Phi3.5, Llama2, Llama3.1 \\
    & Qwen2.5, PMC-Llama, Meditron \\ [2pt]
    Template & w/o RAG, Pre-Q, Aft-Q, Aft-C \\ [2pt]
    Evaluation & Entropy, Best Prob, Accuracy, ACE \\
    \bottomrule
    \end{tabular}
    }
    \caption{Experimental settings used in our research.}
    \label{tab:exp-settings}
\end{table}

\begin{figure*}[t!]
    \centering
    \includegraphics[width=\textwidth]{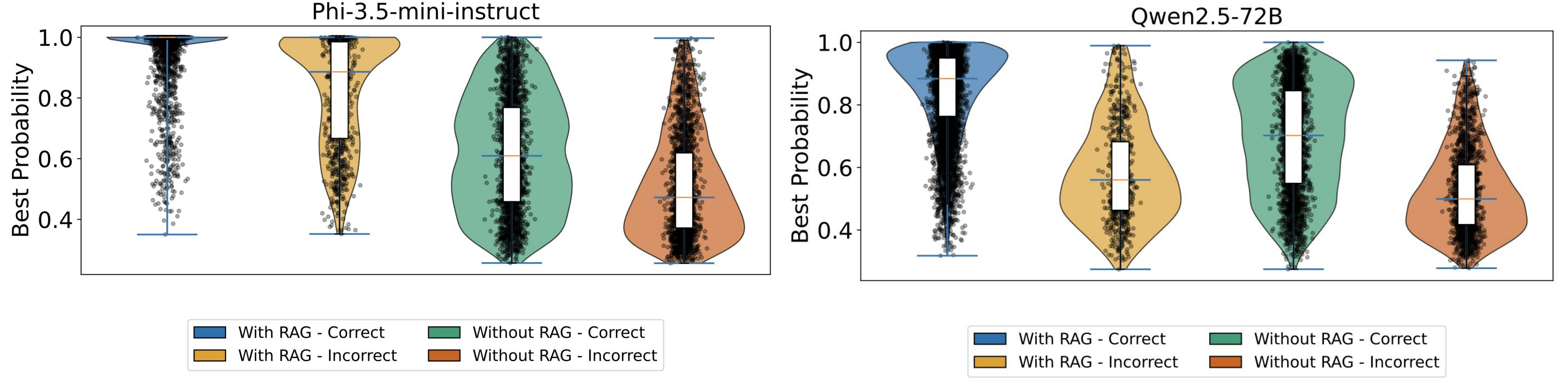}
    \caption{The transition of experimental results using MedMCQA. The figure classifies correctly answered and incorrectly answered questions, illustrating how their distributions shift. This visualization corresponds to the Ans1 setting, with plots for all three conditions: Pre-Q, Aft-Q, and Aft-C.}
    \label{fig:medmcqa-violin}
\end{figure*}

\input{tables/medmcqa-results}

\section{Results}
Table~\ref{tab:medmcqa-results} presents the experimental results using MedMCQA, while Table~\ref{tab:pubmedqa-results} shows the results for PubMedQA.  
When distinguishing between correctly answered and incorrectly answered questions, Phi and Qwen exhibited ideal behavior from an entropy perspective. 
Specifically, inserting supporting documents for the correct answers led to higher entropy, whereas inserting only unrelated documents resulted in lower entropy.  
In contrast, other models, e.g., Llama2, Llama3.1, and Gemma2, produced unexpected results, suggesting that Llama and Gemma may struggle to process inserted documents effectively.  
Furthermore, Qwen and Phi demonstrated the ability to determine whether an inserted document was relevant to the answer, leading to provide strong evidence that they function as suitable generators.

\begin{figure*}[t!]
    \centering
    \includegraphics[width=\textwidth]{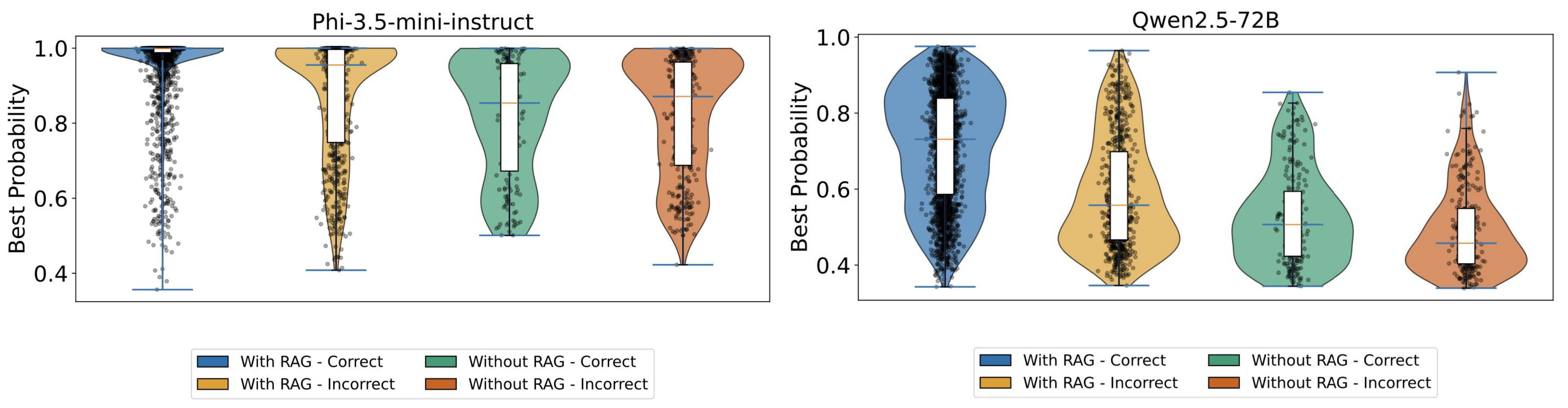}
    \caption{The transition of experimental results using PubMedQA.
    The figure classifies correctly answered and incorrectly answered questions, illustrating how their distributions shift.
    This visualization corresponds to the Ans1 setting, with plots for all three conditions: Pre-Q, Aft-Q, and Aft-C.}
    \label{fig:pubmedqa-violin}
\end{figure*}

\input{tables/pubmedqa-results}

\input{tables/medmcqa-ece-ace-results}

\input{tables/pubmedqa-ece-ace-results}

\section{Analysis \& Discussion}
\label{analysis-dnd-discussion}
\paragraph{How Does RAG Affect Prediction Probabilities?}
Figure~\ref{fig:medmcqa-violin} presents partial experimental results using MedMCQA, while Figure~\ref{fig:pubmedqa-violin} shows results from PubMedQA.
These figures correspond to the Ans1 setting, where all three phases—Pre-Q, Aft-Q, and Aft-C—are plotted.
A detailed analysis focuses on Phi and Qwen, which exhibited ideal behavior.
When RAG was not applied, i.e., evaluating the models' intrinsic accuracy, the output probabilities were evenly distributed across both datasets.  
Furthermore, the results of Phi-3.5 on PubMedQA reveal that the incorrect predictions tend to concentrate at the upper end, i.e., where output probabilities are high. 
This pattern suggests that the model exhibits overconfidence, making incorrect predictions despite assigning high probabilities.  
When solving a QA task under a deliberate setting that includes supporting documents for correct answers (similar to pseudo-RAG), all models (Phi and Qwen) showed improved output probabilities.
This suggests that the models can assess whether retrieved documents contain useful information.

\paragraph{Model Behavior When Inserting Answer-Containing Documents.}
When explicitly inserting documents that contain the correct answers, Phi and Qwen demonstrated ideal behavior.
For instance, from a correct entropy perspective in Table~\ref{tab:medmcqa-results}, Phi had a value of 0.933 under the w/o RAG setting, which dropped to 0.051 after document insertion. 
Similarly, for Qwen2.5 (72B), entropy decreased from 0.819 to 0.444.
This observation indicates that the models can assess whether an inserted document is relevant to the question.
Moreover, when they determine that the document is unnecessary, they attempt to answer using their own knowledge.  
Further evidence supporting this conclusion comes from cases where inserting unrelated documents did not improve accuracy. 
This suggests that the models selectively utilize external information only when it is deemed useful.  

\paragraph{Behavior of Calibration Error.}  
Table~\ref{tab:medmcqa-ece-ace} presents the evaluation result of ACE using MedMCQA, while Table~\ref{tab:pubmedqa-ece-ace} shows the results for PubMedQA.  
A detailed analysis of Llama and Gemma reveals substantial differences in behavior depending on the model.
Notably, even when inserting entirely correct documents (Ans1), Llama3.1 (70B) experiences a drop in accuracy, whereas Llama3.1 (8B) shows improved accuracy even when inserting completely unrelated documents (Oth3). This stark contrast indicates that even within the same model family, behavior can vary largely. 
Moreover, a comparison between Llama2, Meditron, and Llama3.1 shows considerable differences in behavior, ruling out parameter size as the primary cause. 
These findings suggest that while the Llama series performs well under specific instruction formats, it may negatively impact performance in other cases.  
On the other hand, Qwen and Phi exhibit a clear pattern: inserting entirely unrelated documents (Oth3) worsens ACE, while inserting answer-containing documents (Ans1 or Ans1-Oth2) leads to improvements.
This tendency implies that Phi and Qwen possess the ability to assess whether retrieved documents provide useful information.
These results show that analyzing LLM confidence through predicted probabilities effectively reveals the model's ability to identify meaningful documents.

\paragraph{Did ``Lost in the Middle'' Phenomenon Occur?}  
Our study also examined the ``Lost in the Middle'' phenomenon~\cite{liu-etal-2024-lost} by evaluating the impact of document placement within the template across multiple positions (Pre-Q, Aft-Q, and Aft-C).
Focusing on Phi and Qwen, which exhibited expected behavior in terms of entropy and accuracy, an intriguing pattern emerged.
From an entropy perspective, inserting the document after the answer choices yielded the best results, while from an accuracy perspective, placing it before the answer choices was optimal.  
These findings suggest that when prioritizing the reliability of information, entropy should be the primary metric.

\paragraph{Error Analysis.}  
Appendix~\ref{app:error-analysis} presents the results of the error analysis, which examines how the model makes mistakes.  
PubMedQA consists of three answer choices: yes, no, and maybe, allowing us to analyze the types of errors the model makes.  
For Llama3.1 (8B, 70B), PMC-Llama, and Gemma2, the bin colors remain unchanged, indicating that these models do not incorporate arbitrarily inserted supporting documents (Ans1, Ans1-Oth2).  
Meditron, w/o RAG, outputs ``No'' for all incorrect answers.
However, when a document is provided, it changes all responses to ``Yes,'' revealing an extremely sensitive behavior.

\section{Conclusion}  
Our research explored the impact of Retrieval Augmented Generation (RAG) on model confidence in the medical domain where information reliability is crucial.
We found that when models retrieve relevant documents, they not only boost accuracy but also show higher confidence scores. 
In contrast, irrelevant documents have little effect on improving confidence. 
Several models demonstrate the ability to judge if the retrieved documents connect to the correct answer,
indicating a more discerning use of external information than we anticipated.
Our evaluation metrics provide a clear framework for spotting the best generator models within RAG systems. 
The findings reveal that models adjust their output probabilities in response to the quality of the retrieved documents, which opens up new ways to measure and improve model performance. 
These insights help refine RAG methods, making them more reliable for high-stakes applications.

\section{Limitations}
\subsection{The Experiments of the Other Domain}
Our study prioritizes domains where RAG is applied, focusing specifically on the medical domain to analyze confidence. 
To advance further, it becomes necessary to validate RAG in domains such as finance and analyze its confidence in contexts requiring highly reliable information.

\subsection{Further Analyzing New RAG Architecture}
Our study focused exclusively on analyzing the basic RAG architecture. 
While the standard RAG framework directly utilizes retrieved documents within the LLM, newer RAG architectures incorporate various control mechanisms. 
Moving forward, it is essential to analyze these advanced architectures from the perspective of confidence as well.

\subsection{Other Metrics for Evaluation}
The evaluation metrics used in this study, ACE, may have some drawbacks.~\cite{kull2019beyond, kumar2019verified, baan-etal-2022-stop}.
Since LLMs outputs are not always strictly correct or incorrect, researchers often use Prediction Rejection Ratio (PRR), which measures the correlation between confidence scores and output quality.~\cite{fadeeva-etal-2023-lm, vashurin2025benchmarkinguncertaintyquantificationmethods, he-etal-2024-trust, ozaki2025texttiger}. 
However, our study focuses on a multiple-QA task, where each question has a uniquely defined correct answer. 
Additionally, the models were evaluated using force decoding. 
Given these conditions, ACE serves as appropriate evaluation metrics.

\subsection{Methods for Generating Model Outputs}
This study deliberately avoids generating free-text responses from models.
Instead, it retrieves answer candidates using force decoding
This decision stems from an observation in prior research: many studies rely heavily on regular expressions to extract correct answers, leading to substantial accuracy variations even when using the same QA task and model.  
(\url{https://github.com/Teddy-XiongGZ/MedRAG}, \url{https://github.com/epfLLM/meditron}, \url{https://github.com/chaoyi-wu/PMC-LLaMA}.)
To address this issue, we select answer choices based on the model's inherent output probabilities.
This approach avoids introducing dependencies on specific evaluation metrics, which would otherwise occur if the model were required to generate explanations using Chain-of-Thought (CoT) or produce confidence scores.

\section{Ethical Considerations}
\subsection{The Possibility of Dataset Bias}
The datasets and retrieval mechanisms employed in our study may carry inherent biases, which could influence the model's predictions and potentially affect fairness in decision-making. 
Recognizing these biases, we advocate for the use of diverse and representative datasets to minimize their impact. 
Additionally, we uphold transparency by analyzing the interplay between confidence and accuracy, providing users with clearer insights into the system's limitations and confidence. 
However, we emphasize the need for human oversight, as no automated system can guarantee infallibility.

\subsection{AI Assistant Tools}
We used ChatGPT~\footnote{\url{https://chatgpt.com/}} and DeepL~\footnote{\url{https://www.deepl.com/ja/translator}} to translate sentences to English and accelerate our research.

\bibliography{custom, anthology}
\appendix

\section{Example Appendix}
\label{sec:appendix}

\subsection{Detailed Model Settings}
\label{app: detailed_model_setttings}
The PMC-Llama model was quantized to half-precision, while the 70B / 72B models were quantized to 4-bit precision for experimentation.
The implementation relied on the Transformers library~\cite{wolf-etal-2020-transformers} and bitsandbytes~\cite{dettmers2022gpt3}.

\begin{table}[h]
\setlength{\tabcolsep}{2pt}
\resizebox{\columnwidth}{!}{%
    \begin{tabular}{lcl}
        \toprule
        Model  & Params & HuggingFace Name  \\
        \midrule
        Phi-3.5 & \texttt{3.8B} & \texttt{microsoft/Phi-3.5-mini-instruct} \\
        PMC-Llama & \texttt{13B} & \texttt{axiong/PMC\_LLaMA\_13B}  \\
        LLama2 & \texttt{70B}    & \texttt{meta-llama/Llama-2-70b-chat-hf}     \\
        Meditron  & \texttt{70B}  & \texttt{epfl-llm/meditron-70b}  \\
        Llama3.1 & \texttt{8B}  & \texttt{meta-llama/Llama-3.1-8B}  \\
        Llama3.1 & \texttt{70B}  & \texttt{meta-llama/Llama-3.1-70B}  \\
        Gemma2 & \texttt{2B} & \texttt{google/gemma-2-2b}  \\
        Qwen2.5 & \texttt{14B} & \texttt{Qwen/Qwen2.5-14B} \\
        Qwen2.5 & \texttt{72B} & \texttt{Qwen/Qwen2.5-72B} \\
        \bottomrule
    \end{tabular}
    \label{tab:model_settings_with_huggingface}
}
\caption{Detailed name of models.}
\end{table}

\subsection{Dataset Selection}
\label{app: dataset selection}
The dataset selection is based on prior research by~\citet{xiong-etal-2024-benchmarking}. 
From the datasets they used, we select those that include both QA pairs and explanatory passages that justify the answers (MedMCQA and PubMedQA) for this study.

Since the test set for MedMCQA is not publicly available, our study used the dev set as the test set, following the approach adopted in MIRAGE.\footnote{\url{https://huggingface.co/datasets/openlifescienceai/medmcqa}}.
We used the datasets, especially MedMCQA\footnote{\url{https://github.com/MedMCQA/MedMCQA}}, PubMedQA\footnote{\url{https://github.com/pubmedqa/pubmedqa}}.

\subsection{Details of the Input Format}  
As described in Section~\ref{methods}, we determine the selected choice based on the output probabilities assigned by LLMs to the given candidates. 
To prevent answer choices from being split into multiple tokens by the tokenizer, we replace them with A, B, C, and D before feeding them into the model.
This approach ensures a fair comparison across models, even for answer choices that would otherwise span multiple tokens.

\subsection {Inference Settings}
In this study, as far as inference which needs to use GPUs, all experiments were conducted on a single NVIDIA RTX A6000 and NVIDIA GeForce RTX 3090 GPU.

\subsection{Why Do We Focus on the Medical Domain?}
Among the various domains where information reliability is crucial (e.g., finance, law, autonomous driving, and healthcare), we chose to focus on healthcare for the following reasons:
\begin{itemize}
    \item Complexity and Scale of Medical Texts: Medical documents are inherently complex and vast in scope, making them particularly suitable for RAG-based approaches. Combined with the critical importance of information reliability in this field, focusing on healthcare becomes a highly significant choice.
    \item Challenges in Real-World Applications: Questions involving detailed patient information, such as medical histories and symptoms, often overwhelm retrieval systems, making it difficult to identify crucial diagnostic clues. Furthermore, in practical applications, patient conditions and individual characteristics vary widely. Differences in age, medical history, genetic factors, and lifestyle often lead to variations in treatment for the same disease. Providing inaccurate information in such scenarios can result in severe consequences.~\cite{sohn2024rationale}
    \item Established Significance of BioNLP: The prominence of the healthcare domain is evident from the long-standing ``BioNLP'' workshop, which has been held for over two decades.\footnote{\url{https://aclweb.org/aclwiki/BioNLP_Workshop}}
    \item Emerging Trends in Healthcare RAG: Efforts to improve RAG performance in the medical domain have led to developments like Self-BioRAG, which leverages confidence scores. Its popularity and significant citation count highlight this field as a trending area of research. ~\cite{jeong2024improving}
    These points illustrate the rationale behind our focus on the healthcare domain.
\end{itemize}

\subsection{Details of Evaluation Metrics}
\label{app:deetails-of-evaluation-metrics}
\paragraph{Expected Calibration Error (ECE)} \par  
Calibration error metrics evaluate whether a model's predicted probabilities align with actual accuracy in QA tasks.
For instance, if a model assigns a 90\% probability to an answer, the accuracy of such predictions should also be 90\% for optimal calibration.  
Expected Calibration Error (ECE)~\cite{naeini2015obtaining, guo2017calibration} quantifies this discrepancy by segmenting the predicted probability range into multiple bins and computing the difference between the predicted probability and the observed accuracy within each bin as follows:

\begin{equation}
\text{ECE} = \sum_{m=1}^{M} \frac{|B_m|}{n} \left| \text{acc}(B_m) - \text{conf}(B_m) \right|   
\end{equation}

Here, $M$ denotes the number of bins, $B_m$ represents the set of samples within bin $m$, $|B_m|$ is the number of samples in bin $m$, and $n$ is the total number of samples. 
$\text{acc}(B_m)$ refers to the accuracy within bin $B_m$, while $\text{conf}(B_m)$ indicates the average confidence of predictions in bin $m$. 
ECE is computed as the weighted average of the absolute differences between the accuracy and confidence across bins, where the weights correspond to the proportion of samples in each bin.

\paragraph{Adaptive Calibration Error (ACE)} \par  
ACE performs binning so that the number of samples in each bin remains constant. This approach ensures a more stable evaluation within each bin:
\begin{equation}
    \text{ACE} = \frac{1}{KR} \sum_{k=1}^{K} \sum_{r=1}^{R} \left| \text{acc}(r, k) - \text{conf}(r, k) \right|
\end{equation}
Here, $K$ denotes the number of classes, $R$ represents the number of bins, $\text{acc}(r, k)$ indicates the accuracy in bin $r$ for class $k$, and $\text{conf}(r, k)$ denotes the confidence of predictions in the same bin and class.

\subsection{The results using Expected Calibration Error (ECE)}
The results using ECE are presented in Table~\ref{tab:medmcqa-ece-only} and Table~\ref{tab:pubmedqa-ece-only}.  
As discussed in Section~\ref{evaluation-metrics}, ~\citet{proskurina-etal-2024-quantization} and ~\citet{ulmer-etal-2022-exploring} have pointed out that ACE is a more suitable calibration error metric for multi-class classification problems, while ECE is better suited for binary classification.
Nevertheless, we include ECE results for completeness and additional verification.

\subsection{Violin plot}
\begin{figure}[h!]
    \centering
    \begin{subfigure}[b]{\linewidth}
        \includegraphics[width=\linewidth]{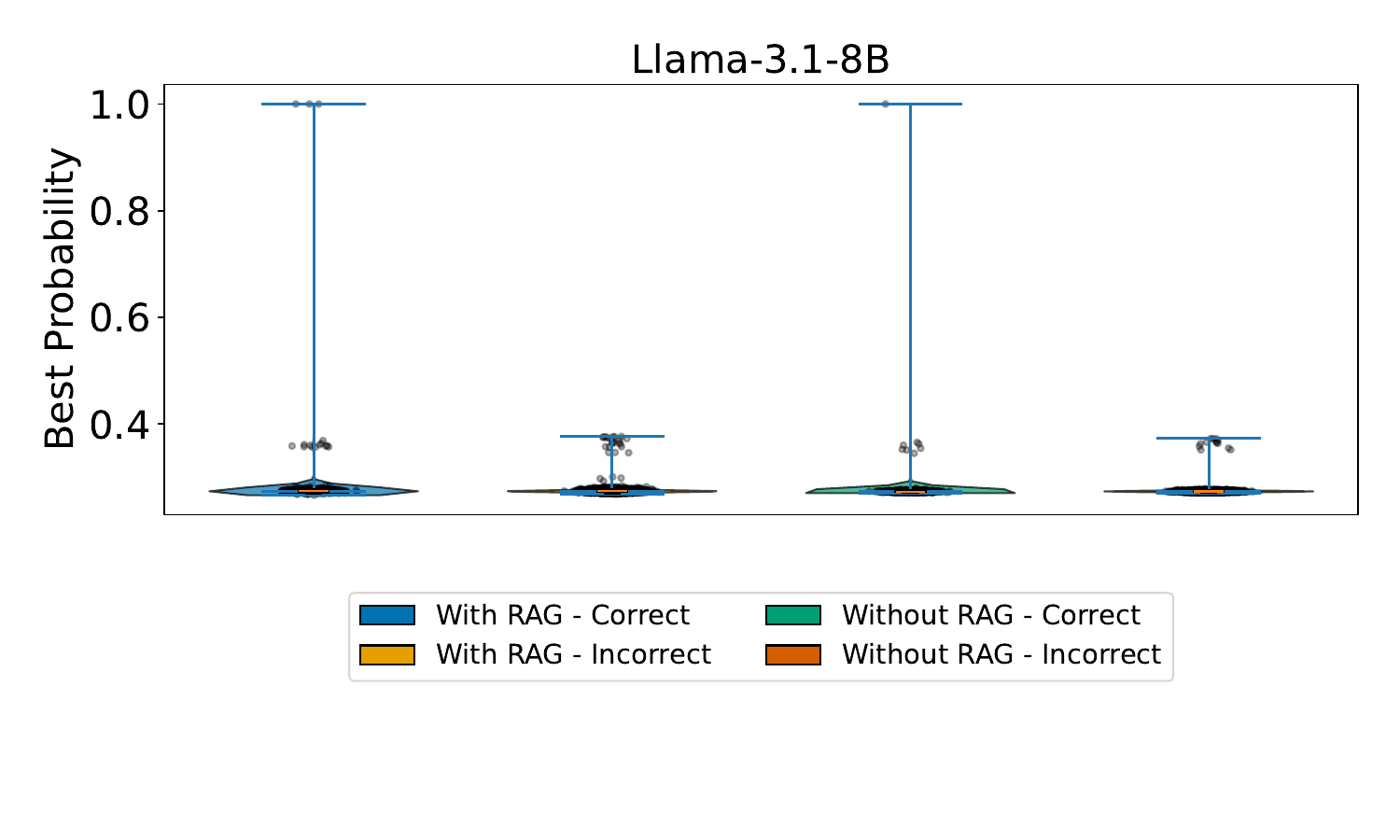}
        \caption{The result of Llama3.1 (8B) using MedMCQA.}
        \label{fig:medmcqa-llama318-violin}
    \end{subfigure}
    \begin{subfigure}[b]{\linewidth}
        \includegraphics[width=\linewidth]{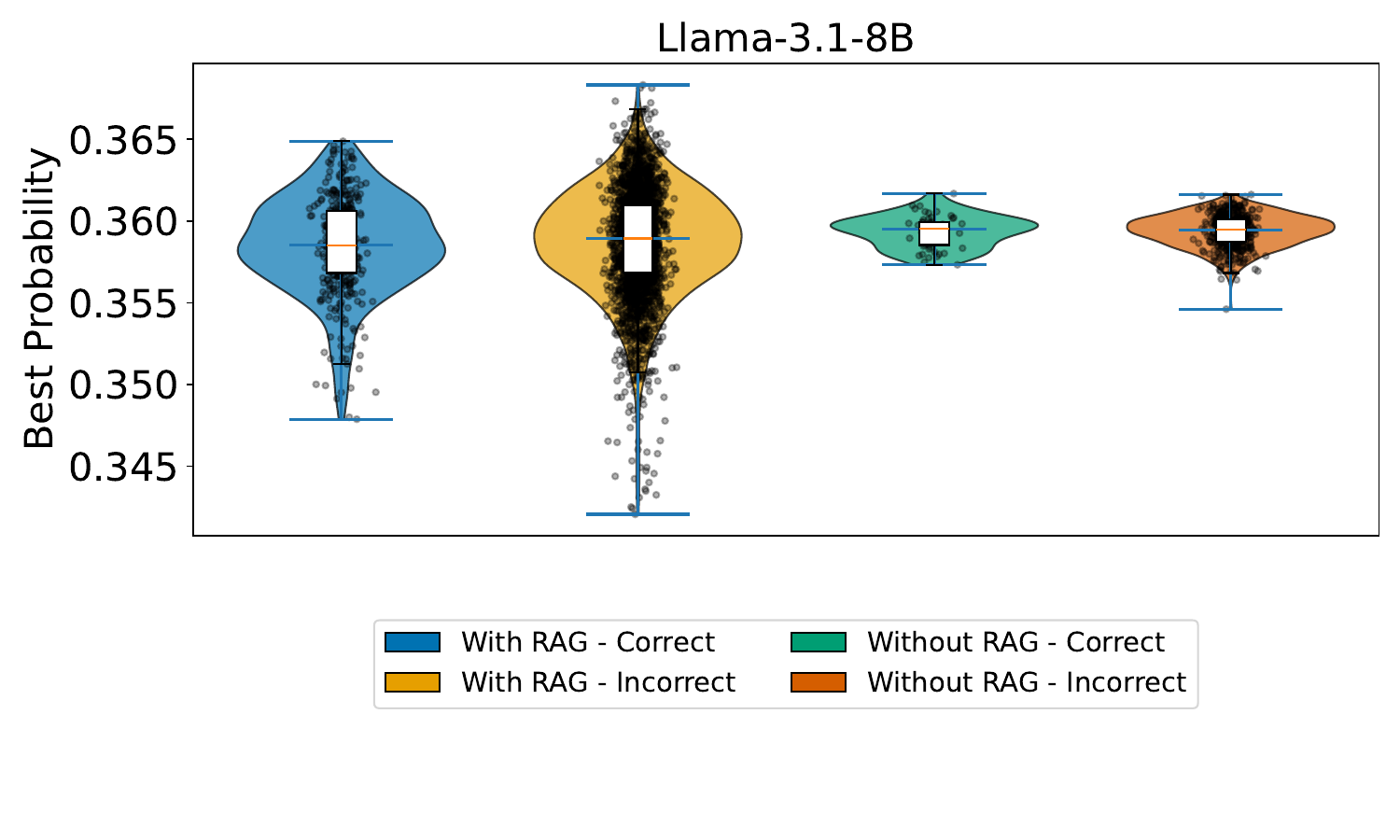}
        \caption{The result of Llama3.1 (8B) using PubMedQA.}
        \label{fig:pubmedqa-llama318-violin}
    \end{subfigure}
    \begin{subfigure}[b]{\linewidth}
        \includegraphics[width=\linewidth]{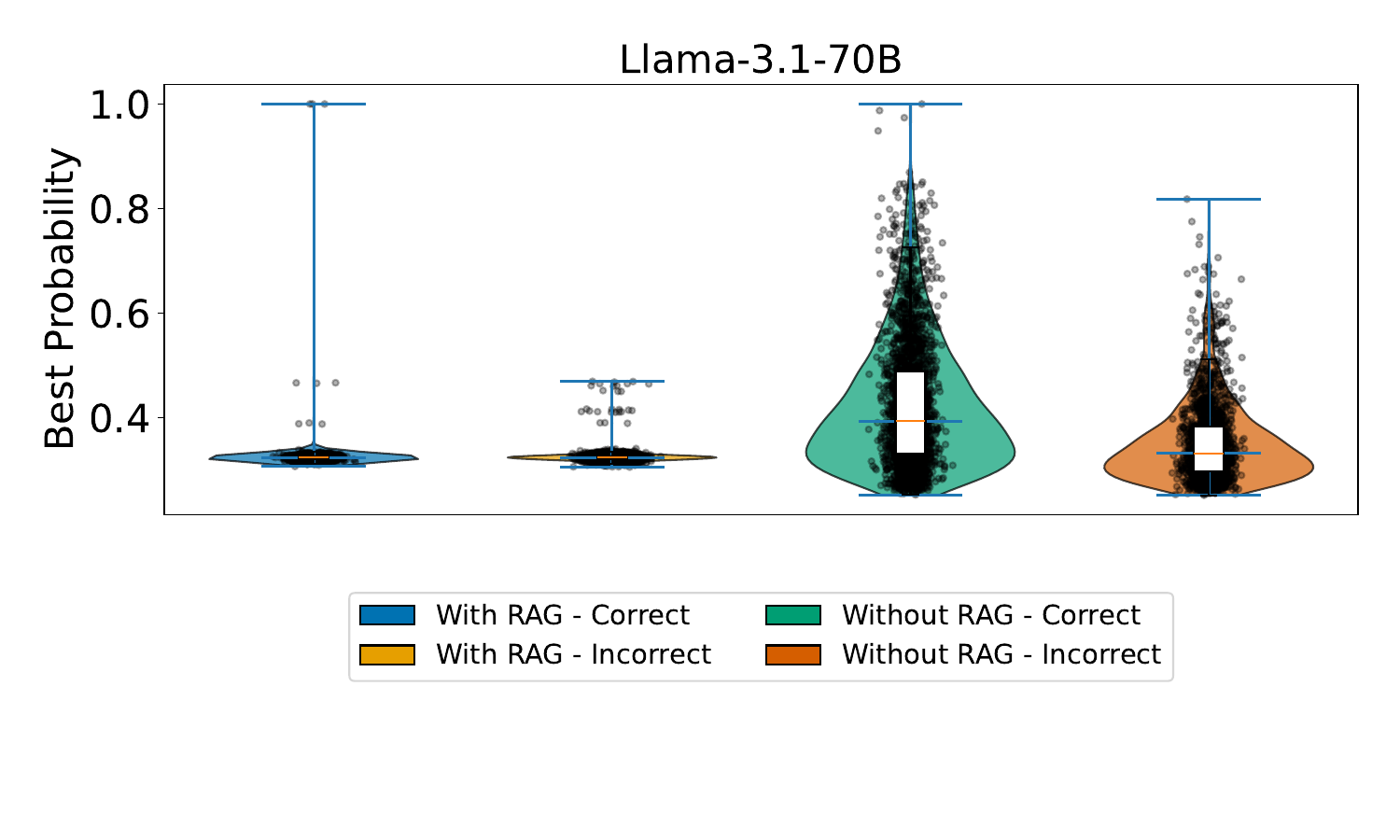}
        \caption{The result of Llama3.1 (70B) using MedMCQA.}
        \label{fig:medmcqa-llama3170-violin}
    \end{subfigure}
    \begin{subfigure}[b]{\linewidth}
        \includegraphics[width=\linewidth]{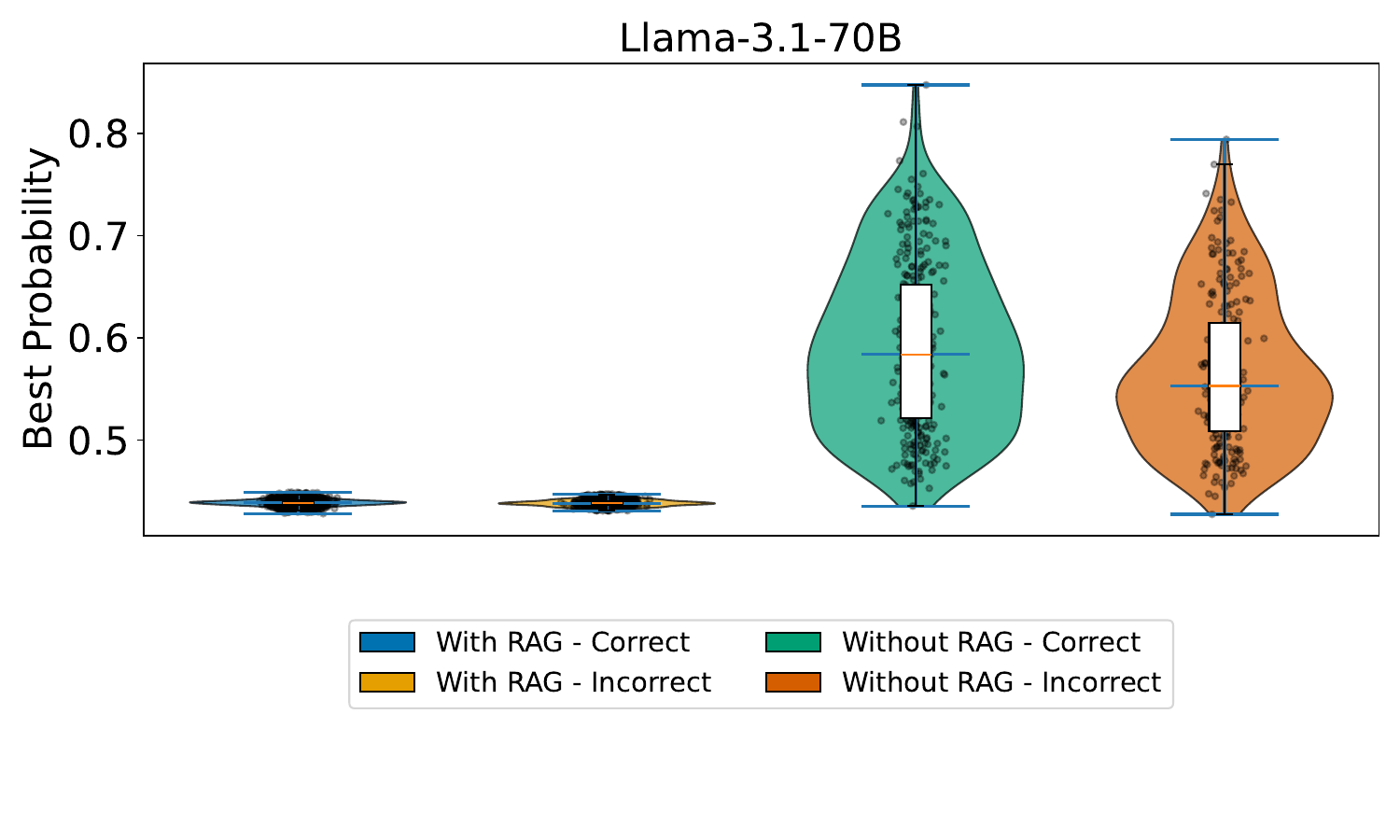}
        \caption{The result of Llama3.1 (70B) using PubMedQA.}
        \label{fig:pubmedqa-llama3170-violin}
    \end{subfigure}
    \caption{Results from MedMCQA and PubMedQA using Llama3.1.}
    \label{fig:all-results}
\end{figure}

Figures~\ref{fig:medmcqa-llama318-violin} and \ref{fig:pubmedqa-llama318-violin} present the violin plot results for Llama3.1 (8B), while Figures~\ref{fig:medmcqa-llama3170-violin} and \ref{fig:pubmedqa-llama3170-violin} show the results for Llama3.1 (70B).  
The Llama models exhibit notably low output probabilities for candidate answer choices when no supporting documents are inserted. 
Furthermore, even when explicitly inserting documents containing supporting evidence, the output probabilities do not improve significantly. 
This suggests that these models may strictly adhere to predefined instructions and struggle to incorporate additional contextual information.

\section{Future Direction}
In this study, we used a dataset containing correct answer choices along with supporting rationale passages for QA tasks.
In the future, it may be possible to focus on non-medical domains by drawing on previous work that semi-automatically generates questions using LLMs~\cite{ozaki2024bqa, sakai-etal-2024-mcsqa}.
There are also studies on explanation generation~\cite{ozaki-etal-2025-towards, hayashi-etal-2024-towards}, which could inform the generation of supporting rationales.

\input{tables/medmcqa-ece-results}

\input{tables/pubmedqa-ece-results}

\clearpage
\onecolumn
\subsection{Error Analysis}
\label{app:error-analysis}
\begin{figure}[h!]
    \centering
    \begin{subfigure}[b]{\linewidth}
        \includegraphics[width=\linewidth]{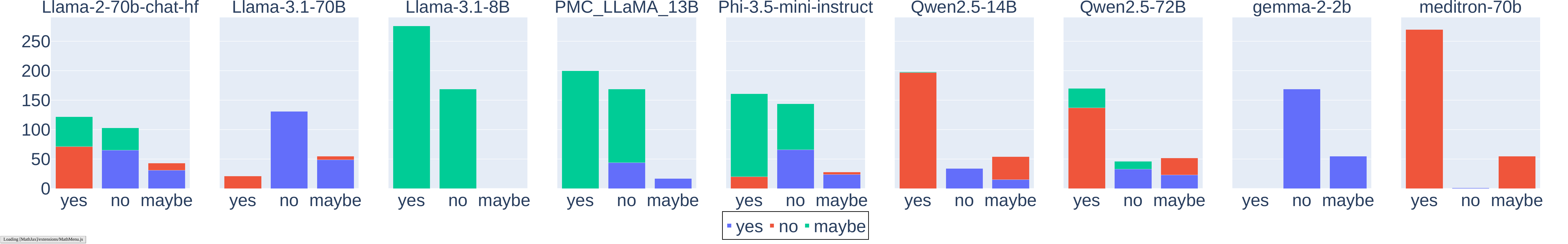}
        \caption{When not inserting anything (w/o RAG).}
        \label{fig:pubmedqa-base-ea}
    \end{subfigure}
    \begin{subfigure}[b]{\linewidth}
        \includegraphics[width=\linewidth]{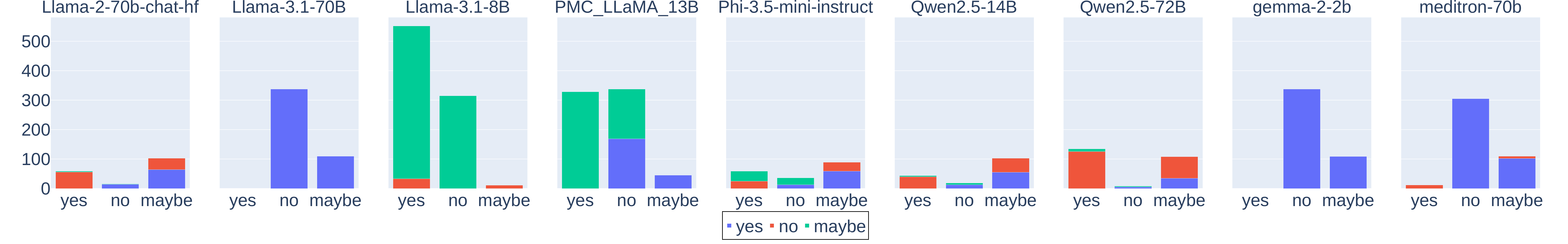}
        \caption{When inserting a document containing the correct answer (Ans1).}
        \label{fig:pubmedqa-ans1-ea}
    \end{subfigure}
    \begin{subfigure}[b]{\linewidth}
        \includegraphics[width=\linewidth]{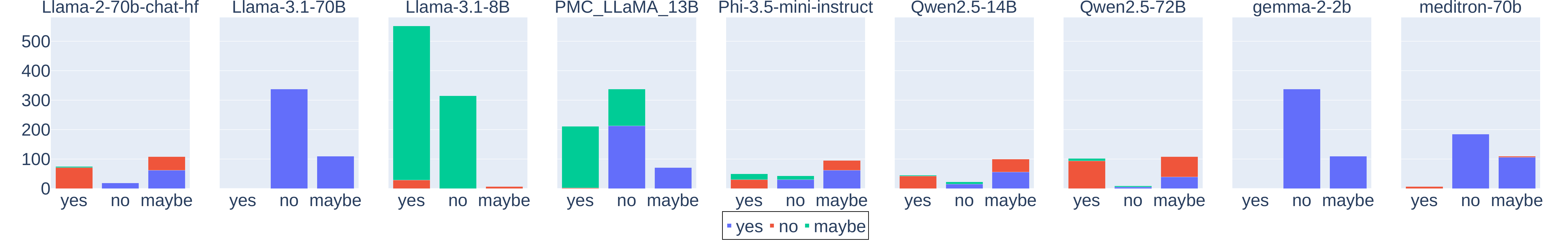}
        \caption{When inserting one relevant document containing the correct answer and two unrelated documents (Ans1-Oth2).}
        \label{fig:epubmedqa-ans1-oth2-ea}
    \end{subfigure}
    \begin{subfigure}[b]{\linewidth}
        \includegraphics[width=\linewidth]{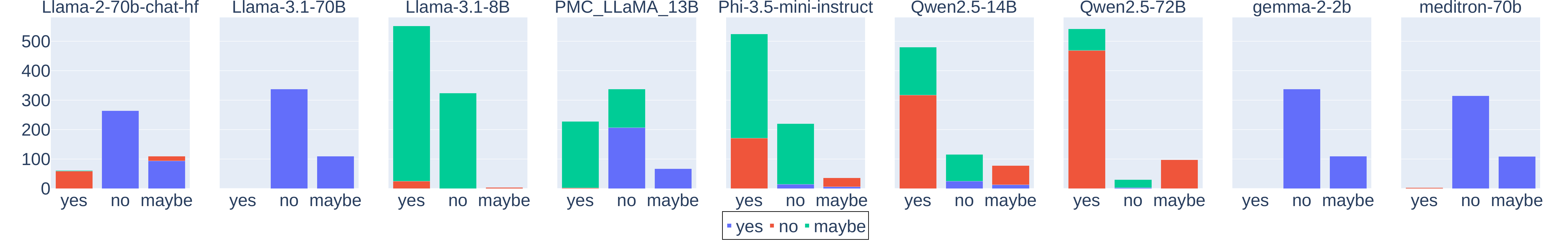}
        \caption{When inserting three unrelated documents (Oth3).}
        \label{fig:pubmedqa-oth3-ea}
    \end{subfigure}
    \caption{
    Error analysis on PubMedQA: This figure illustrates how the model misclassified answers in relation to the correct ones.
We gather the questions the model got wrong, group the items that actually had the correct answer (gold answer) into bins, and use colors to show how the model erred.    
}
    \label{fig:all-results-ea}
\end{figure}

Figure~\ref{fig:all-results-ea} presents a plot illustrating the types of errors made on incorrectly answered questions.  
In PubMedQA, the answer choices consist of three options: yes, no, and maybe, allowing for detailed error analysis.
Each bin represents the gold answer, and the plot visualizes the distribution of incorrect predictions for each question.
The colors within the plot indicate how the model misclassified the answers.

\onecolumn
\clearpage
\subsection{Prompts}
Below are examples of prompts with and without RAG. 
When RAG is applied, three patterns-Pre-Question, After-Question, and After-Choice—are used in our study.

\begin{tcolorbox}[title=Prompt without RAG, boxrule=1pt]
\small
\texttt{You are a helpful medical expert, and your task is to answer a multi-choice medical question using the relevant documents. Please first think step-by-step and then choose the answer from the provided options. Your responses will be used for research purposes only, so please have a definite answer.} \\

\texttt{Here is the question:} \\
\{\texttt{question}\} \\

\texttt{Here are the potential choices:} \\
\{\texttt{choice0}\} \\
\{\texttt{choice1}\} \\
\{\texttt{choice2}\} \\
\{\texttt{choice3}\} \\

\texttt{Answer:}
\end{tcolorbox}

\begin{tcolorbox}[title=Prompt with RAG, boxrule=1pt]
\small
\blue{\texttt{Here are the relevant documents: \textbf{(Pre-Question)}}} \\
\blue{\{\texttt{context}\}} \\ 

\texttt{You are a helpful medical expert, and your task is to answer a multi-choice medical question using the relevant documents. Please first think step-by-step and then choose the answer from the provided options. Your responses will be used for research purposes only, so please have a definite answer.} \\

\blue{\texttt{Here are the relevant documents: \textbf{(After-Question)}}} \\
\blue{\{\texttt{context}\}} \\ 

\texttt{Here is the question:} \\
\{\texttt{question}\} \\

\blue{\texttt{Here are the relevant documents: \textbf{(After-Choice)}}} \\
\blue{\{\texttt{context}\}} \\ 

\texttt{Here are the potential choices:} \\
\{\texttt{choice0}\} \\
\{\texttt{choice1}\} \\
\{\texttt{choice2}\} \\
\{\texttt{choice3}\} \\

\texttt{Answer:}
\end{tcolorbox}

\end{document}

%% file: tables/medmcqa-results.tex
\begin{table*}[t!]
    \resizebox{\textwidth}{!}{%
    \centering
    \begin{tabular}{llcccccccccccccccc}
        \toprule
        \multicolumn{18}{c}{\multirow{2}{*}{\textbf{MedMCQA (Entropy and Best Probability)}}} \\
        & \\
        \midrule
        \multirow{2.5}{*}{\textbf{Model}} & \multirow{2.5}{*}{\textbf{Pattern}}  & \multicolumn{4}{c}{\textbf{Entropy (Correct) $\downarrow$}} & \multicolumn{4}{c}{\textbf{Best Prob (Correct) $\uparrow$}} & \multicolumn{4}{c}{\textbf{Entropy (Incorrect) $\downarrow$}} & \multicolumn{4}{c}{\textbf{Best Prob (Incorrect) $\downarrow$}} \\
        \cmidrule(lr){3-6} \cmidrule(lr){7-10} \cmidrule(lr){11-14} \cmidrule(lr){15-18}
        & & \textbf{None} & \textbf{Ans1} & \textbf{Ans1-Oth2} & \textbf{Oth3} & \textbf{None} & \textbf{Ans1} & \textbf{Ans1-Oth2} & \textbf{Oth3} & \textbf{None} & \textbf{Ans1} & \textbf{Ans1-Oth2} & \textbf{Oth3} & \textbf{None} & \textbf{Ans1} & \textbf{Ans1-Oth2} & \textbf{Oth3} \\
        \midrule
\multirow{5}{*}{Llama-2-70b-chat-hf} & w/o RAG & 1.24$_{\pm 1.11}$ & -- & -- & -- & 0.42$_{\pm 0.55}$ & -- & -- & -- & 1.28$_{\pm 1.27}$ & -- & -- & -- & 0.38$_{\pm 0.38}$ & -- & -- & -- \\
\cmidrule{2-18}
 & Pre-Q & -- & 1.12$_{\pm 0.16}$ & 1.12$_{\pm 0.16}$ & 1.27$_{\pm 0.10}$ & -- & 0.54$_{\pm 0.12}$ & 0.54$_{\pm 0.12}$ & 0.41$_{\pm 0.09}$ & -- & 1.27$_{\pm 0.08}$ & 1.28$_{\pm 0.08}$ & 1.31$_{\pm 0.06}$ & -- & 0.41$_{\pm 0.07}$ & 0.40$_{\pm 0.08}$ & \textbf{0.38$_{\pm 0.07}$} \\
 & Aft-Q & -- & \textbf{1.11$_{\pm 0.16}$} & 1.15$_{\pm 0.16}$ & 1.29$_{\pm 0.10}$ & -- & \textbf{0.55$_{\pm 0.12}$} & 0.52$_{\pm 0.13}$ & 0.40$_{\pm 0.09}$ & -- & \textbf{1.27$_{\pm 0.09}$} & 1.29$_{\pm 0.08}$ & 1.31$_{\pm 0.06}$ & -- & 0.41$_{\pm 0.08}$ & 0.40$_{\pm 0.08}$ & 0.38$_{\pm 0.07}$ \\
 & Aft-C & -- & 1.15$_{\pm 0.16}$ & 1.23$_{\pm 0.12}$ & 1.30$_{\pm 0.10}$ & -- & 0.51$_{\pm 0.13}$ & 0.46$_{\pm 0.11}$ & 0.39$_{\pm 0.09}$ & -- & 1.28$_{\pm 0.09}$ & 1.31$_{\pm 0.07}$ & 1.31$_{\pm 0.07}$ & -- & 0.40$_{\pm 0.08}$ & 0.38$_{\pm 0.08}$ & 0.38$_{\pm 0.08}$ \\
\midrule
\multirow{5}{*}{Llama-3.1-70B} & w/o RAG & 1.24$_{\pm 1.24}$ & -- & -- & -- & 0.42$_{\pm 0.42}$ & -- & -- & -- & 1.31$_{\pm 1.31}$ & -- & -- & -- & 0.35$_{\pm 0.32}$ & -- & -- & -- \\
\cmidrule{2-18}
 & Pre-Q & -- & 1.35$_{\pm 0.07}$ & 1.35$_{\pm 0.07}$ & 1.35$_{\pm 0.07}$ & -- & 0.32$_{\pm 0.03}$ & 0.33$_{\pm 0.03}$ & 0.33$_{\pm 0.03}$ & -- & 1.35$_{\pm 0.02}$ & 1.35$_{\pm 0.02}$ & 1.35$_{\pm 0.02}$ & -- & \textbf{0.32$_{\pm 0.01}$} & 0.32$_{\pm 0.01}$ & 0.33$_{\pm 0.01}$ \\
 & Aft-Q & -- & 1.35$_{\pm 0.07}$ & 1.35$_{\pm 0.07}$ & 1.35$_{\pm 0.07}$ & -- & 0.33$_{\pm 0.03}$ & 0.33$_{\pm 0.03}$ & 0.33$_{\pm 0.03}$ & -- & 1.35$_{\pm 0.02}$ & 1.35$_{\pm 0.02}$ & 1.35$_{\pm 0.02}$ & -- & 0.32$_{\pm 0.01}$ & 0.33$_{\pm 0.01}$ & 0.33$_{\pm 0.01}$ \\
 & Aft-C & -- & 1.35$_{\pm 0.07}$ & 1.35$_{\pm 0.07}$ & 1.35$_{\pm 0.07}$ & -- & 0.33$_{\pm 0.03}$ & 0.33$_{\pm 0.03}$ & 0.33$_{\pm 0.03}$ & -- & 1.35$_{\pm 0.02}$ & 1.35$_{\pm 0.02}$ & 1.35$_{\pm 0.02}$ & -- & 0.32$_{\pm 0.01}$ & 0.33$_{\pm 0.01}$ & 0.33$_{\pm 0.01}$ \\
\midrule
\multirow{5}{*}{Llama-3.1-8B} & w/o RAG & 1.38$_{\pm 1.38}$ & -- & -- & -- & 0.28$_{\pm 0.28}$ & -- & -- & -- & 1.38$_{\pm 1.38}$ & -- & -- & -- & 0.27$_{\pm 0.27}$ & -- & -- & -- \\
\cmidrule{2-18}
 & Pre-Q & -- & 1.38$_{\pm 0.07}$ & \textbf{1.38$_{\pm 0.07}$} & 1.38$_{\pm 0.07}$ & -- & 0.28$_{\pm 0.03}$ & 0.28$_{\pm 0.03}$ & 0.28$_{\pm 0.03}$ & -- & 1.38$_{\pm 0.02}$ & 1.38$_{\pm 0.02}$ & 1.38$_{\pm 0.02}$ & -- & 0.27$_{\pm 0.01}$ & 0.27$_{\pm 0.01}$ & 0.27$_{\pm 0.01}$ \\
 & Aft-Q & -- & 1.38$_{\pm 0.07}$ & 1.38$_{\pm 0.07}$ & 1.38$_{\pm 0.06}$ & -- & 0.28$_{\pm 0.03}$ & 0.28$_{\pm 0.03}$ & 0.28$_{\pm 0.03}$ & -- & 1.38$_{\pm 0.02}$ & \textbf{1.38$_{\pm 0.02}$} & 1.38$_{\pm 0.02}$ & -- & 0.27$_{\pm 0.01}$ & 0.27$_{\pm 0.01}$ & 0.27$_{\pm 0.01}$ \\
 & Aft-C & -- & 1.38$_{\pm 0.07}$ & 1.38$_{\pm 0.07}$ & 1.38$_{\pm 0.07}$ & -- & 0.28$_{\pm 0.03}$ & \textbf{0.28$_{\pm 0.03}$} & 0.28$_{\pm 0.03}$ & -- & 1.38$_{\pm 0.02}$ & 1.38$_{\pm 0.02}$ & 1.38$_{\pm 0.02}$ & -- & 0.27$_{\pm 0.01}$ & 0.27$_{\pm 0.01}$ & 0.27$_{\pm 0.01}$ \\
\midrule
\multirow{5}{*}{meditron-70b} & w/o RAG & 1.23$_{\pm 1.05}$ & -- & -- & -- & 0.42$_{\pm 0.57}$ & -- & -- & -- & 1.25$_{\pm 1.19}$ & -- & -- & -- & 0.40$_{\pm 0.40}$ & -- & -- & -- \\
\cmidrule{2-18}
 & Pre-Q & -- & 1.11$_{\pm 0.17}$ & \textbf{1.05$_{\pm 0.19}$} & 1.23$_{\pm 0.11}$ & -- & 0.54$_{\pm 0.13}$ & \textbf{0.57$_{\pm 0.14}$} & 0.47$_{\pm 0.09}$ & -- & 1.24$_{\pm 0.10}$ & \textbf{1.19$_{\pm 0.11}$} & 1.27$_{\pm 0.08}$ & -- & 0.43$_{\pm 0.08}$ & 0.47$_{\pm 0.08}$ & 0.43$_{\pm 0.08}$ \\
 & Aft-Q & -- & 1.10$_{\pm 0.17}$ & 1.09$_{\pm 0.17}$ & 1.21$_{\pm 0.11}$ & -- & 0.54$_{\pm 0.13}$ & 0.55$_{\pm 0.13}$ & 0.48$_{\pm 0.09}$ & -- & 1.23$_{\pm 0.10}$ & 1.20$_{\pm 0.10}$ & 1.23$_{\pm 0.10}$ & -- & 0.44$_{\pm 0.08}$ & 0.47$_{\pm 0.08}$ & 0.47$_{\pm 0.09}$ \\
 & Aft-C & -- & 1.15$_{\pm 0.17}$ & 1.28$_{\pm 0.10}$ & 1.30$_{\pm 0.08}$ & -- & 0.51$_{\pm 0.13}$ & 0.43$_{\pm 0.09}$ & 0.41$_{\pm 0.07}$ & -- & 1.27$_{\pm 0.08}$ & 1.31$_{\pm 0.06}$ & 1.31$_{\pm 0.06}$ & -- & 0.41$_{\pm 0.07}$ & 0.40$_{\pm 0.07}$ & 0.41$_{\pm 0.07}$ \\
\midrule
\multirow{5}{*}{PMC-LLaMA-13B} & w/o RAG & 1.00$_{\pm 1.00}$ & -- & -- & -- & 0.56$_{\pm 0.56}$ & -- & -- & -- & 1.05$_{\pm 1.05}$ & -- & -- & -- & 0.53$_{\pm 0.31}$ & -- & -- & -- \\
\cmidrule{2-18}
 & Pre-Q & -- & 1.36$_{\pm 0.06}$ & 1.36$_{\pm 0.06}$ & 1.36$_{\pm 0.06}$ & -- & 0.33$_{\pm 0.05}$ & 0.32$_{\pm 0.05}$ & 0.32$_{\pm 0.05}$ & -- & 1.37$_{\pm 0.03}$ & 1.37$_{\pm 0.02}$ & 1.37$_{\pm 0.02}$ & -- & 0.31$_{\pm 0.04}$ & \textbf{0.31$_{\pm 0.03}$} & 0.31$_{\pm 0.03}$ \\
 & Aft-Q & -- & 1.36$_{\pm 0.06}$ & 1.36$_{\pm 0.06}$ & 1.36$_{\pm 0.06}$ & -- & 0.33$_{\pm 0.05}$ & 0.32$_{\pm 0.05}$ & 0.31$_{\pm 0.05}$ & -- & 1.36$_{\pm 0.03}$ & 1.37$_{\pm 0.03}$ & 1.37$_{\pm 0.03}$ & -- & 0.32$_{\pm 0.04}$ & 0.31$_{\pm 0.03}$ & 0.31$_{\pm 0.04}$ \\
 & Aft-C & -- & 1.36$_{\pm 0.06}$ & 1.35$_{\pm 0.07}$ & 1.36$_{\pm 0.07}$ & -- & 0.33$_{\pm 0.05}$ & 0.33$_{\pm 0.05}$ & 0.33$_{\pm 0.05}$ & -- & 1.36$_{\pm 0.03}$ & 1.36$_{\pm 0.03}$ & 1.36$_{\pm 0.03}$ & -- & 0.32$_{\pm 0.04}$ & 0.32$_{\pm 0.04}$ & 0.32$_{\pm 0.04}$ \\
\midrule
\multirow{5}{*}{Gemma-2-2b} & w/o RAG & 1.17$_{\pm 1.11}$ & -- & -- & -- & 0.52$_{\pm 0.56}$ & -- & -- & -- & 1.18$_{\pm 1.13}$ & -- & -- & -- & 0.52$_{\pm 0.51}$ & -- & -- & -- \\
\cmidrule{2-18}
 & Pre-Q & -- & 1.12$_{\pm 0.08}$ & 1.13$_{\pm 0.07}$ & 1.15$_{\pm 0.06}$ & -- & 0.55$_{\pm 0.05}$ & 0.54$_{\pm 0.04}$ & 0.52$_{\pm 0.04}$ & -- & 1.17$_{\pm 0.05}$ & 1.16$_{\pm 0.04}$ & 1.16$_{\pm 0.04}$ & -- & \textbf{0.51$_{\pm 0.04}$} & 0.51$_{\pm 0.04}$ & 0.52$_{\pm 0.03}$ \\
 & Aft-Q & -- & 1.13$_{\pm 0.07}$ & 1.14$_{\pm 0.06}$ & 1.15$_{\pm 0.06}$ & -- & 0.55$_{\pm 0.05}$ & 0.53$_{\pm 0.04}$ & 0.52$_{\pm 0.04}$ & -- & 1.17$_{\pm 0.05}$ & 1.16$_{\pm 0.04}$ & 1.15$_{\pm 0.04}$ & -- & 0.51$_{\pm 0.04}$ & 0.51$_{\pm 0.04}$ & 0.52$_{\pm 0.03}$ \\
 & Aft-C & -- & \textbf{1.11$_{\pm 0.08}$} & 1.12$_{\pm 0.07}$ & 1.13$_{\pm 0.06}$ & -- & \textbf{0.56$_{\pm 0.05}$} & 0.55$_{\pm 0.05}$ & 0.54$_{\pm 0.04}$ & -- & 1.16$_{\pm 0.05}$ & 1.14$_{\pm 0.05}$ & \textbf{1.13$_{\pm 0.05}$} & -- & 0.53$_{\pm 0.04}$ & 0.54$_{\pm 0.04}$ & 0.54$_{\pm 0.04}$ \\
\midrule
\midrule
\multirow{5}{*}{Phi-3.5} & w/o RAG & 0.93$_{\pm 0.05}$ & -- & -- & -- & 0.62$_{\pm 0.98}$ & -- & -- & -- & 1.09$_{\pm 0.39}$ & -- & -- & -- & 0.51$_{\pm 0.51}$ & -- & -- & -- \\
\cmidrule{2-18}
 & Pre-Q & -- & 0.06$_{\pm 0.17}$ & 0.07$_{\pm 0.18}$ & 0.24$_{\pm 0.32}$ & -- & 0.98$_{\pm 0.08}$ & 0.98$_{\pm 0.08}$ & 0.90$_{\pm 0.15}$ & -- & \textbf{0.39$_{\pm 0.34}$} & 0.43$_{\pm 0.35}$ & 0.49$_{\pm 0.38}$ & -- & 0.84$_{\pm 0.18}$ & 0.82$_{\pm 0.18}$ & 0.80$_{\pm 0.19}$ \\
 & Aft-Q & -- & \textbf{0.05$_{\pm 0.16}$} & 0.07$_{\pm 0.18}$ & 0.34$_{\pm 0.36}$ & -- & \textbf{0.98$_{\pm 0.07}$} & 0.97$_{\pm 0.08}$ & 0.87$_{\pm 0.17}$ & -- & 0.45$_{\pm 0.35}$ & 0.46$_{\pm 0.35}$ & 0.50$_{\pm 0.37}$ & -- & 0.81$_{\pm 0.18}$ & 0.81$_{\pm 0.18}$ & 0.80$_{\pm 0.19}$ \\
 & Aft-C & -- & 0.09$_{\pm 0.19}$ & 0.14$_{\pm 0.22}$ & 0.27$_{\pm 0.32}$ & -- & 0.97$_{\pm 0.09}$ & 0.95$_{\pm 0.10}$ & 0.90$_{\pm 0.15}$ & -- & 0.45$_{\pm 0.34}$ & 0.44$_{\pm 0.36}$ & 0.42$_{\pm 0.35}$ & -- & 0.81$_{\pm 0.18}$ & 0.82$_{\pm 0.19}$ & 0.84$_{\pm 0.17}$ \\
\midrule
\multirow{5}{*}{Qwen2.5-14B} & w/o RAG & 0.86$_{\pm 0.48}$ & -- & -- & -- & 0.67$_{\pm 0.85}$ & -- & -- & -- & 1.06$_{\pm 1.03}$ & -- & -- & -- & 0.55$_{\pm 0.49}$ & -- & -- & -- \\
\cmidrule{2-18}
 & Pre-Q & -- & 0.52$_{\pm 0.33}$ & 0.53$_{\pm 0.35}$ & 0.89$_{\pm 0.30}$ & -- & 0.84$_{\pm 0.15}$ & 0.83$_{\pm 0.16}$ & 0.65$_{\pm 0.18}$ & -- & \textbf{1.03$_{\pm 0.23}$} & 1.05$_{\pm 0.23}$ & 1.07$_{\pm 0.22}$ & -- & 0.56$_{\pm 0.15}$ & 0.55$_{\pm 0.15}$ & 0.54$_{\pm 0.15}$ \\
 & Aft-Q & -- & \textbf{0.48$_{\pm 0.32}$} & 0.51$_{\pm 0.33}$ & 0.92$_{\pm 0.29}$ & -- & \textbf{0.85$_{\pm 0.14}$} & 0.84$_{\pm 0.16}$ & 0.63$_{\pm 0.17}$ & -- & 1.04$_{\pm 0.24}$ & 1.05$_{\pm 0.23}$ & 1.06$_{\pm 0.21}$ & -- & 0.56$_{\pm 0.15}$ & 0.55$_{\pm 0.15}$ & 0.55$_{\pm 0.14}$ \\
 & Aft-C & -- & 0.66$_{\pm 0.35}$ & 0.80$_{\pm 0.32}$ & 1.00$_{\pm 0.24}$ & -- & 0.77$_{\pm 0.17}$ & 0.71$_{\pm 0.18}$ & 0.59$_{\pm 0.16}$ & -- & 1.12$_{\pm 0.20}$ & 1.12$_{\pm 0.20}$ & 1.15$_{\pm 0.17}$ & -- & 0.51$_{\pm 0.13}$ & 0.51$_{\pm 0.14}$ & \textbf{0.49$_{\pm 0.12}$} \\
\midrule
\multirow{5}{*}{Qwen2.5-72B} & w/o RAG & 0.82$_{\pm 0.44}$ & -- & -- & -- & 0.69$_{\pm 0.86}$ & -- & -- & -- & 1.09$_{\pm 0.87}$ & -- & -- & -- & 0.53$_{\pm 0.53}$ & -- & -- & -- \\
\cmidrule{2-18}
 & Pre-Q & -- & 0.52$_{\pm 0.31}$ & 0.56$_{\pm 0.32}$ & 0.75$_{\pm 0.32}$ & -- & 0.83$_{\pm 0.14}$ & 0.80$_{\pm 0.15}$ & 0.71$_{\pm 0.18}$ & -- & 0.91$_{\pm 0.26}$ & \textbf{0.87$_{\pm 0.24}$} & 0.94$_{\pm 0.24}$ & -- & 0.61$_{\pm 0.16}$ & 0.63$_{\pm 0.15}$ & 0.62$_{\pm 0.15}$ \\
 & Aft-Q & -- & \textbf{0.44$_{\pm 0.30}$} & 0.54$_{\pm 0.31}$ & 0.83$_{\pm 0.31}$ & -- & \textbf{0.86$_{\pm 0.13}$} & 0.82$_{\pm 0.15}$ & 0.68$_{\pm 0.17}$ & -- & 0.99$_{\pm 0.27}$ & 0.97$_{\pm 0.24}$ & 0.99$_{\pm 0.22}$ & -- & 0.57$_{\pm 0.16}$ & 0.59$_{\pm 0.15}$ & 0.60$_{\pm 0.15}$ \\
 & Aft-C & -- & 0.53$_{\pm 0.32}$ & 0.65$_{\pm 0.32}$ & 0.85$_{\pm 0.28}$ & -- & 0.82$_{\pm 0.15}$ & 0.77$_{\pm 0.16}$ & 0.67$_{\pm 0.17}$ & -- & 1.00$_{\pm 0.25}$ & 1.02$_{\pm 0.21}$ & 0.99$_{\pm 0.22}$ & -- & 0.57$_{\pm 0.15}$ & 0.57$_{\pm 0.14}$ & 0.61$_{\pm 0.14}$ \\
\bottomrule
    \end{tabular}
    }
    \caption{
    Experimental result on MedMCQA.
    \textbf{Bold} indicates the best value among the models. 
    Specifically, the lowest entropy and the highest best probability (Correct case) are highlighted.
    This table has numerical values and their standard deviations.
    }
    \label{tab:medmcqa-results}
\end{table*}

%% file: tables/pubmedqa-results.tex
\begin{table*}[t!]
    \resizebox{\textwidth}{!}{%
    \centering
    \begin{tabular}{llcccccccccccccccccccc}
        \toprule
        \multicolumn{18}{c}{\multirow{2}{*}{\textbf{PubMedQA (Entropy and Best Probability)}}} \\
        & \\
        \midrule
        \multirow{2.5}{*}{\textbf{Model}} & \multirow{2.5}{*}{\textbf{Pattern}}  & \multicolumn{4}{c}{\textbf{Entropy (Correct) $\downarrow$}} & \multicolumn{4}{c}{\textbf{Best Prob (Correct) $\uparrow$}} & \multicolumn{4}{c}{\textbf{Entropy (Incorrect) $\downarrow$}} & \multicolumn{4}{c}{\textbf{Best Prob (Incorrect) $\downarrow$}} \\
        \cmidrule(lr){3-6} \cmidrule(lr){7-10} \cmidrule(lr){11-14} \cmidrule(lr){15-18}
        & & \textbf{None} & \textbf{Ans1} & \textbf{Ans1-Oth2} & \textbf{Oth3} & \textbf{None} & \textbf{Ans1} & \textbf{Ans1-Oth2} & \textbf{Oth3} & \textbf{None} & \textbf{Ans1} & \textbf{Ans1-Oth2} & \textbf{Oth3} & \textbf{None} & \textbf{Ans1} & \textbf{Ans1-Oth2} & \textbf{Oth3} \\
        \midrule
\multirow{5}{*}{Llama-2-70b-chat-hf} & w/o RAG & 0.83$_{\pm 0.83}$ & -- & -- & -- & 0.62$_{\pm 0.62}$ & -- & -- & -- & 0.93$_{\pm 0.93}$ & -- & -- & -- & 0.55$_{\pm 0.38}$ & -- & -- & -- \\
\cmidrule{2-18}
 & Pre-Q & -- & 1.12$_{\pm 0.16}$ & 1.12$_{\pm 0.16}$ & 1.27$_{\pm 0.10}$ & -- & 0.54$_{\pm 0.12}$ & 0.54$_{\pm 0.12}$ & 0.41$_{\pm 0.09}$ & -- & 1.27$_{\pm 0.08}$ & 1.28$_{\pm 0.08}$ & 1.31$_{\pm 0.06}$ & -- & 0.41$_{\pm 0.07}$ & 0.40$_{\pm 0.08}$ & \textbf{0.38$_{\pm 0.07}$} \\
 & Aft-Q & -- & 1.11$_{\pm 0.16}$ & 1.15$_{\pm 0.16}$ & 1.29$_{\pm 0.10}$ & -- & 0.55$_{\pm 0.12}$ & 0.52$_{\pm 0.13}$ & 0.40$_{\pm 0.09}$ & -- & 1.27$_{\pm 0.09}$ & 1.29$_{\pm 0.08}$ & 1.31$_{\pm 0.06}$ & -- & 0.41$_{\pm 0.08}$ & 0.40$_{\pm 0.08}$ & 0.38$_{\pm 0.07}$ \\
 & Aft-C & -- & 1.15$_{\pm 0.16}$ & 1.23$_{\pm 0.12}$ & 1.30$_{\pm 0.10}$ & -- & 0.51$_{\pm 0.13}$ & 0.46$_{\pm 0.11}$ & 0.39$_{\pm 0.09}$ & -- & 1.28$_{\pm 0.09}$ & 1.31$_{\pm 0.07}$ & 1.31$_{\pm 0.07}$ & -- & 0.40$_{\pm 0.08}$ & 0.38$_{\pm 0.08}$ & 0.38$_{\pm 0.08}$ \\
\midrule
\multirow{5}{*}{Llama-3.1-70B} & w/o RAG & 0.86$_{\pm 0.86}$ & -- & -- & -- & 0.59$_{\pm 0.59}$ & -- & -- & -- & 0.87$_{\pm 0.87}$ & -- & -- & -- & 0.57$_{\pm 0.32}$ & -- & -- & -- \\
\cmidrule{2-18}
 & Pre-Q & -- & 1.35$_{\pm 0.07}$ & 1.35$_{\pm 0.07}$ & 1.35$_{\pm 0.07}$ & -- & 0.32$_{\pm 0.03}$ & 0.33$_{\pm 0.03}$ & 0.33$_{\pm 0.03}$ & -- & 1.35$_{\pm 0.02}$ & 1.35$_{\pm 0.02}$ & 1.35$_{\pm 0.02}$ & -- & \textbf{0.32$_{\pm 0.01}$} & 0.32$_{\pm 0.01}$ & 0.33$_{\pm 0.01}$ \\
 & Aft-Q & -- & 1.35$_{\pm 0.07}$ & 1.35$_{\pm 0.07}$ & 1.35$_{\pm 0.07}$ & -- & 0.33$_{\pm 0.03}$ & 0.33$_{\pm 0.03}$ & 0.33$_{\pm 0.03}$ & -- & 1.35$_{\pm 0.02}$ & 1.35$_{\pm 0.02}$ & 1.35$_{\pm 0.02}$ & -- & 0.32$_{\pm 0.01}$ & 0.33$_{\pm 0.01}$ & 0.33$_{\pm 0.01}$ \\
 & Aft-C & -- & 1.35$_{\pm 0.07}$ & 1.35$_{\pm 0.07}$ & 1.35$_{\pm 0.07}$ & -- & 0.33$_{\pm 0.03}$ & 0.33$_{\pm 0.03}$ & 0.33$_{\pm 0.03}$ & -- & 1.35$_{\pm 0.02}$ & 1.35$_{\pm 0.02}$ & 1.35$_{\pm 0.02}$ & -- & 0.32$_{\pm 0.01}$ & 0.33$_{\pm 0.01}$ & 0.33$_{\pm 0.01}$ \\
\midrule
\multirow{5}{*}{Llama-3.1-8B} & w/o RAG & 1.09$_{\pm 1.09}$ & -- & -- & -- & 0.36$_{\pm 0.36}$ & -- & -- & -- & 1.09$_{\pm 1.09}$ & -- & -- & -- & 0.36$_{\pm 0.27}$ & -- & -- & -- \\
\cmidrule{2-18}
 & Pre-Q & -- & 1.38$_{\pm 0.07}$ & 1.38$_{\pm 0.07}$ & 1.38$_{\pm 0.07}$ & -- & 0.28$_{\pm 0.03}$ & 0.28$_{\pm 0.03}$ & 0.28$_{\pm 0.03}$ & -- & 1.38$_{\pm 0.02}$ & 1.38$_{\pm 0.02}$ & 1.38$_{\pm 0.02}$ & -- & 0.27$_{\pm 0.01}$ & 0.27$_{\pm 0.01}$ & \textbf{0.27$_{\pm 0.01}$} \\
 & Aft-Q & -- & 1.38$_{\pm 0.07}$ & 1.38$_{\pm 0.07}$ & 1.38$_{\pm 0.06}$ & -- & 0.28$_{\pm 0.03}$ & 0.28$_{\pm 0.03}$ & 0.28$_{\pm 0.03}$ & -- & 1.38$_{\pm 0.02}$ & 1.38$_{\pm 0.02}$ & 1.38$_{\pm 0.02}$ & -- & 0.27$_{\pm 0.01}$ & 0.27$_{\pm 0.01}$ & 0.27$_{\pm 0.01}$ \\
 & Aft-C & -- & 1.38$_{\pm 0.07}$ & 1.38$_{\pm 0.07}$ & 1.38$_{\pm 0.07}$ & -- & 0.28$_{\pm 0.03}$ & 0.28$_{\pm 0.03}$ & 0.28$_{\pm 0.03}$ & -- & 1.38$_{\pm 0.02}$ & 1.38$_{\pm 0.02}$ & 1.38$_{\pm 0.02}$ & -- & 0.27$_{\pm 0.01}$ & 0.27$_{\pm 0.01}$ & 0.27$_{\pm 0.01}$ \\
\midrule
\multirow{5}{*}{meditron-70b} & w/o RAG & 0.94$_{\pm 0.94}$ & -- & -- & -- & 0.54$_{\pm 0.57}$ & -- & -- & -- & 0.94$_{\pm 0.94}$ & -- & -- & -- & 0.52$_{\pm 0.40}$ & -- & -- & -- \\
\cmidrule{2-18}
 & Pre-Q & -- & 1.11$_{\pm 0.17}$ & 1.05$_{\pm 0.19}$ & 1.23$_{\pm 0.11}$ & -- & 0.54$_{\pm 0.13}$ & \textbf{0.57$_{\pm 0.14}$} & 0.47$_{\pm 0.09}$ & -- & 1.24$_{\pm 0.10}$ & 1.19$_{\pm 0.11}$ & 1.27$_{\pm 0.08}$ & -- & 0.43$_{\pm 0.08}$ & 0.47$_{\pm 0.08}$ & 0.43$_{\pm 0.08}$ \\
 & Aft-Q & -- & 1.10$_{\pm 0.17}$ & 1.09$_{\pm 0.17}$ & 1.21$_{\pm 0.11}$ & -- & 0.54$_{\pm 0.13}$ & 0.55$_{\pm 0.13}$ & 0.48$_{\pm 0.09}$ & -- & 1.23$_{\pm 0.10}$ & 1.20$_{\pm 0.10}$ & 1.23$_{\pm 0.10}$ & -- & 0.44$_{\pm 0.08}$ & 0.47$_{\pm 0.08}$ & 0.47$_{\pm 0.09}$ \\
 & Aft-C & -- & 1.15$_{\pm 0.17}$ & 1.28$_{\pm 0.10}$ & 1.30$_{\pm 0.08}$ & -- & 0.51$_{\pm 0.13}$ & 0.43$_{\pm 0.09}$ & 0.41$_{\pm 0.07}$ & -- & 1.27$_{\pm 0.08}$ & 1.31$_{\pm 0.06}$ & 1.31$_{\pm 0.06}$ & -- & 0.41$_{\pm 0.07}$ & \textbf{0.40$_{\pm 0.07}$} & 0.41$_{\pm 0.07}$ \\
\midrule
\multirow{5}{*}{PMC-LLaMA-13B} & w/o RAG & 1.08$_{\pm 1.08}$ & -- & -- & -- & 0.40$_{\pm 0.40}$ & -- & -- & -- & 1.08$_{\pm 1.08}$ & -- & -- & -- & 0.40$_{\pm 0.31}$ & -- & -- & -- \\
\cmidrule{2-18}
 & Pre-Q & -- & 1.36$_{\pm 0.06}$ & 1.36$_{\pm 0.06}$ & 1.36$_{\pm 0.06}$ & -- & 0.33$_{\pm 0.05}$ & 0.32$_{\pm 0.05}$ & 0.32$_{\pm 0.05}$ & -- & 1.37$_{\pm 0.03}$ & 1.37$_{\pm 0.02}$ & 1.37$_{\pm 0.02}$ & -- & 0.31$_{\pm 0.04}$ & \textbf{0.31$_{\pm 0.03}$} & 0.31$_{\pm 0.03}$ \\
 & Aft-Q & -- & 1.36$_{\pm 0.06}$ & 1.36$_{\pm 0.06}$ & 1.36$_{\pm 0.06}$ & -- & 0.33$_{\pm 0.05}$ & 0.32$_{\pm 0.05}$ & 0.31$_{\pm 0.05}$ & -- & 1.36$_{\pm 0.03}$ & 1.37$_{\pm 0.03}$ & 1.37$_{\pm 0.03}$ & -- & 0.32$_{\pm 0.04}$ & 0.31$_{\pm 0.03}$ & 0.31$_{\pm 0.04}$ \\
 & Aft-C & -- & 1.36$_{\pm 0.06}$ & 1.35$_{\pm 0.07}$ & 1.36$_{\pm 0.07}$ & -- & 0.33$_{\pm 0.05}$ & 0.33$_{\pm 0.05}$ & 0.33$_{\pm 0.05}$ & -- & 1.36$_{\pm 0.03}$ & 1.36$_{\pm 0.03}$ & 1.36$_{\pm 0.03}$ & -- & 0.32$_{\pm 0.04}$ & 0.32$_{\pm 0.04}$ & 0.32$_{\pm 0.04}$ \\
\midrule
\multirow{5}{*}{Gemma-2-2b} & w/o RAG & 0.93$_{\pm 0.93}$ & -- & -- & -- & 0.61$_{\pm 0.61}$ & -- & -- & -- & 0.93$_{\pm 0.93}$ & -- & -- & -- & 0.61$_{\pm 0.51}$ & -- & -- & -- \\
\cmidrule{2-18}
 & Pre-Q & -- & 1.12$_{\pm 0.08}$ & 1.13$_{\pm 0.07}$ & 1.15$_{\pm 0.06}$ & -- & 0.55$_{\pm 0.05}$ & 0.54$_{\pm 0.04}$ & 0.52$_{\pm 0.04}$ & -- & 1.17$_{\pm 0.05}$ & 1.16$_{\pm 0.04}$ & 1.16$_{\pm 0.04}$ & -- & \textbf{0.51$_{\pm 0.04}$} & 0.51$_{\pm 0.04}$ & 0.52$_{\pm 0.03}$ \\
 & Aft-Q & -- & 1.13$_{\pm 0.07}$ & 1.14$_{\pm 0.06}$ & 1.15$_{\pm 0.06}$ & -- & 0.55$_{\pm 0.05}$ & 0.53$_{\pm 0.04}$ & 0.52$_{\pm 0.04}$ & -- & 1.17$_{\pm 0.05}$ & 1.16$_{\pm 0.04}$ & 1.15$_{\pm 0.04}$ & -- & 0.51$_{\pm 0.04}$ & 0.51$_{\pm 0.04}$ & 0.52$_{\pm 0.03}$ \\
 & Aft-C & -- & 1.11$_{\pm 0.08}$ & 1.12$_{\pm 0.07}$ & 1.13$_{\pm 0.06}$ & -- & 0.56$_{\pm 0.05}$ & 0.55$_{\pm 0.05}$ & 0.54$_{\pm 0.04}$ & -- & 1.16$_{\pm 0.05}$ & 1.14$_{\pm 0.05}$ & 1.13$_{\pm 0.05}$ & -- & 0.53$_{\pm 0.04}$ & 0.54$_{\pm 0.04}$ & 0.54$_{\pm 0.04}$ \\
\midrule
\midrule
\multirow{5}{*}{Phi-3.5} & w/o RAG & 0.40$_{\pm 0.05}$ & -- & -- & -- & 0.81$_{\pm 0.98}$ & -- & -- & -- & 0.41$_{\pm 0.39}$ & -- & -- & -- & 0.82$_{\pm 0.80}$ & -- & -- & -- \\
\cmidrule{2-18}
 & Pre-Q & -- & 0.06$_{\pm 0.17}$ & 0.07$_{\pm 0.18}$ & 0.24$_{\pm 0.32}$ & -- & 0.98$_{\pm 0.08}$ & 0.98$_{\pm 0.08}$ & 0.90$_{\pm 0.15}$ & -- & \textbf{0.39$_{\pm 0.34}$} & 0.43$_{\pm 0.35}$ & 0.49$_{\pm 0.38}$ & -- & 0.84$_{\pm 0.18}$ & 0.82$_{\pm 0.18}$ & 0.80$_{\pm 0.19}$ \\
 & Aft-Q & -- & \textbf{0.05$_{\pm 0.16}$} & 0.07$_{\pm 0.18}$ & 0.34$_{\pm 0.36}$ & -- & \textbf{0.98$_{\pm 0.07}$} & 0.97$_{\pm 0.08}$ & 0.87$_{\pm 0.17}$ & -- & 0.45$_{\pm 0.35}$ & 0.46$_{\pm 0.35}$ & 0.50$_{\pm 0.37}$ & -- & 0.81$_{\pm 0.18}$ & 0.81$_{\pm 0.18}$ & \textbf{0.80$_{\pm 0.19}$} \\
 & Aft-C & -- & 0.09$_{\pm 0.19}$ & 0.14$_{\pm 0.22}$ & 0.27$_{\pm 0.32}$ & -- & 0.97$_{\pm 0.09}$ & 0.95$_{\pm 0.10}$ & 0.90$_{\pm 0.15}$ & -- & 0.45$_{\pm 0.34}$ & 0.44$_{\pm 0.36}$ & 0.42$_{\pm 0.35}$ & -- & 0.81$_{\pm 0.18}$ & 0.82$_{\pm 0.19}$ & 0.84$_{\pm 0.17}$ \\
\midrule
\multirow{5}{*}{Qwen2.5-14B} & w/o RAG & 0.90$_{\pm 0.48}$ & -- & -- & -- & 0.59$_{\pm 0.85}$ & -- & -- & -- & 0.92$_{\pm 0.92}$ & -- & -- & -- & 0.58$_{\pm 0.49}$ & -- & -- & -- \\
\cmidrule{2-18}
 & Pre-Q & -- & 0.52$_{\pm 0.33}$ & 0.53$_{\pm 0.35}$ & 0.89$_{\pm 0.30}$ & -- & 0.84$_{\pm 0.15}$ & 0.83$_{\pm 0.16}$ & 0.65$_{\pm 0.18}$ & -- & 1.03$_{\pm 0.23}$ & 1.05$_{\pm 0.23}$ & 1.07$_{\pm 0.22}$ & -- & 0.56$_{\pm 0.15}$ & 0.55$_{\pm 0.15}$ & 0.54$_{\pm 0.15}$ \\
 & Aft-Q & -- & \textbf{0.48$_{\pm 0.32}$} & 0.51$_{\pm 0.33}$ & 0.92$_{\pm 0.29}$ & -- & \textbf{0.85$_{\pm 0.14}$} & 0.84$_{\pm 0.16}$ & 0.63$_{\pm 0.17}$ & -- & 1.04$_{\pm 0.24}$ & 1.05$_{\pm 0.23}$ & 1.06$_{\pm 0.21}$ & -- & 0.56$_{\pm 0.15}$ & 0.55$_{\pm 0.15}$ & 0.55$_{\pm 0.14}$ \\
 & Aft-C & -- & 0.66$_{\pm 0.35}$ & 0.80$_{\pm 0.32}$ & 1.00$_{\pm 0.24}$ & -- & 0.77$_{\pm 0.17}$ & 0.71$_{\pm 0.18}$ & 0.59$_{\pm 0.16}$ & -- & 1.12$_{\pm 0.20}$ & 1.12$_{\pm 0.20}$ & 1.15$_{\pm 0.17}$ & -- & 0.51$_{\pm 0.13}$ & 0.51$_{\pm 0.14}$ & \textbf{0.49$_{\pm 0.12}$} \\
\midrule
\multirow{5}{*}{Qwen2.5-72B} & w/o RAG & 0.97$_{\pm 0.44}$ & -- & -- & -- & 0.53$_{\pm 0.86}$ & -- & -- & -- & 1.00$_{\pm 0.87}$ & -- & -- & -- & 0.49$_{\pm 0.49}$ & -- & -- & -- \\
\cmidrule{2-18}
 & Pre-Q & -- & 0.52$_{\pm 0.31}$ & 0.56$_{\pm 0.32}$ & 0.75$_{\pm 0.32}$ & -- & 0.83$_{\pm 0.14}$ & 0.80$_{\pm 0.15}$ & 0.71$_{\pm 0.18}$ & -- & 0.91$_{\pm 0.26}$ & \textbf{0.87$_{\pm 0.24}$} & 0.94$_{\pm 0.24}$ & -- & 0.61$_{\pm 0.16}$ & 0.63$_{\pm 0.15}$ & 0.62$_{\pm 0.15}$ \\
 & Aft-Q & -- & \textbf{0.44$_{\pm 0.30}$} & 0.54$_{\pm 0.31}$ & 0.83$_{\pm 0.31}$ & -- & \textbf{0.86$_{\pm 0.13}$} & 0.82$_{\pm 0.15}$ & 0.68$_{\pm 0.17}$ & -- & 0.99$_{\pm 0.27}$ & 0.97$_{\pm 0.24}$ & 0.99$_{\pm 0.22}$ & -- & 0.57$_{\pm 0.16}$ & 0.59$_{\pm 0.15}$ & 0.60$_{\pm 0.15}$ \\
 & Aft-C & -- & 0.53$_{\pm 0.32}$ & 0.65$_{\pm 0.32}$ & 0.85$_{\pm 0.28}$ & -- & 0.82$_{\pm 0.15}$ & 0.77$_{\pm 0.16}$ & 0.67$_{\pm 0.17}$ & -- & 1.00$_{\pm 0.25}$ & 1.02$_{\pm 0.21}$ & 0.99$_{\pm 0.22}$ & -- & 0.57$_{\pm 0.15}$ & 0.57$_{\pm 0.14}$ & 0.61$_{\pm 0.14}$ \\
\bottomrule
    \end{tabular}
    }
    \caption{
    Experimental results using PubMedQA.
    \textbf{Bold} indicates the best value among the models. 
    Specifically, the lowest entropy and the highest best probability (Correct case) are highlighted.
    This table has numerical values and their standard deviations.
    }
    \label{tab:pubmedqa-results}
\end{table*}

%% file: tables/medmcqa-ece-ace-results.tex
\begin{table*}[t!]
    \centering
    \small
    \resizebox{0.98\textwidth}{!}{
    \renewcommand{\arraystretch}{0.1} 
    \begin{tabular}{cc cccc cccc}
    \toprule
    \multirow{2}{*}{\textbf{Model}} & \multirow{2}{*}{\textbf{Pattern}} & \multicolumn{4}{c}{\textbf{ACE $\downarrow$}} & \multicolumn{4}{c}{\textbf{Accuracy $\uparrow$}} \\
    \cmidrule(lr){3-6}    \cmidrule(lr){7-10}
    & & \textbf{None} & \textbf{Ans1} & \textbf{Ans1-Oth2} & \textbf{Oth3} & \textbf{None} & \textbf{Ans1} & \textbf{Ans1-Oth2} & \textbf{Oth3}  \\
    \midrule
\multirow{40}{*}{Llama2 (70B)} & w/o RAG & 2.208 & -- & -- & -- & 38.322 & -- & -- & --\\
\cmidrule(lr){2-10}
 & Pre-Q & -- & \bluebox{22.359} & \bluebox{25.113} & \bluebox{7.181} & -- & \redbox{72.575} & \redbox{75.340} & \bluebox{32.094} \\  
 & Aft-Q & -- & \bluebox{23.912} & \bluebox{21.132} & \bluebox{10.781} & -- & \redbox{75.567} & \redbox{69.628} & \bluebox{28.105} \\  
 & Aft-C & -- & \bluebox{19.653} & \bluebox{17.514} & \bluebox{9.803} & -- & \redbox{67.498} & \redbox{60.743} & \bluebox{28.876} \\  
\midrule
\multirow{40}{*}{Llama3.1 (70B)} & w/o RAG & 19.582 & -- & -- & -- & 58.977 & -- & -- & --\\
\cmidrule(lr){2-10}
 & Pre-Q & -- & \redbox{11.496} & \redbox{11.580} & \redbox{11.658} & -- & \bluebox{20.898} & \bluebox{20.943} & \bluebox{20.943} \\  
 & Aft-Q & -- & \redbox{11.518} & \redbox{11.671} & \redbox{11.714} & -- & \bluebox{20.898} & \bluebox{20.898} & \bluebox{20.943} \\  
 & Aft-C & -- & \redbox{11.504} & \redbox{11.707} & \redbox{11.795} & -- & \bluebox{20.943} & \bluebox{20.943} & \bluebox{20.943} \\  
\midrule
\multirow{40}{*}{Llama3.1 (8B)} & w/o RAG & 5.423 & -- & -- & -- & 22.209 & -- & -- & --\\
\cmidrule(lr){2-10}
 & Pre-Q & -- & \redbox{4.701} & \redbox{4.254} & \redbox{3.644} & -- & \redbox{23.345} & \redbox{23.209} & \redbox{23.799} \\  
 & Aft-Q & -- & \redbox{4.473} & \redbox{4.632} & \redbox{3.892} & -- & \redbox{23.028} & \redbox{23.209} & \redbox{24.025} \\  
 & Aft-C & -- & \redbox{4.476} & \redbox{4.746} & \redbox{4.990} & -- & \redbox{23.209} & \redbox{23.255} & \redbox{23.663} \\  
\midrule
\multirow{40}{*}{Meditron (70B)} & w/o RAG & 6.412 & -- & -- & -- & 35.525 & -- & -- & --\\
\cmidrule(lr){2-10}
 & Pre-Q & -- & \bluebox{17.684} & \bluebox{7.652} & \bluebox{8.665} & -- & \redbox{67.724} & \redbox{54.034} & \redbox{36.038} \\  
 & Aft-Q & -- & \bluebox{15.894} & \bluebox{9.467} & \bluebox{15.334} & -- & \redbox{66.682} & \redbox{47.144} & \bluebox{31.958} \\  
 & Aft-C & -- & \bluebox{15.101} & \bluebox{6.946} & \bluebox{9.006} & -- & \redbox{62.829} & \bluebox{34.180} & \bluebox{31.913} \\  
\midrule
\multirow{40}{*}{PMC-Llama (13B)} & w/o RAG & 15.671 & -- & -- & -- & 38.107 & -- & -- & --\\
\cmidrule(lr){2-10}
 & Pre-Q & -- & \redbox{4.943} & \redbox{4.367} & \redbox{4.357} & -- & \bluebox{32.729} & \bluebox{31.641} & \bluebox{26.972} \\  
 & Aft-Q & -- & \redbox{4.003} & \redbox{2.550} & \redbox{5.032} & -- & \bluebox{32.910} & \bluebox{30.009} & \bluebox{26.972} \\  
 & Aft-C & -- & \redbox{3.496} & \redbox{3.780} & \redbox{4.397} & -- & \bluebox{33.454} & \bluebox{28.740} & \bluebox{28.060} \\  
\midrule
\multirow{40}{*}{Gemma2 (2B)} & w/o RAG & 19.568 & -- & -- & -- & 32.297 & -- & -- & --\\
\cmidrule(lr){2-10}
 & Pre-Q & -- & \bluebox{25.511} & \bluebox{23.160} & \bluebox{20.520} & -- & \bluebox{31.233} & \bluebox{31.278} & \bluebox{31.278} \\  
 & Aft-Q & -- & \bluebox{24.160} & \bluebox{21.072} & \bluebox{20.618} & -- & \bluebox{31.188} & \bluebox{31.278} & \bluebox{31.278} \\  
 & Aft-C & -- & \bluebox{24.814} & \bluebox{23.118} & \bluebox{22.916} & -- & \bluebox{31.324} & \bluebox{31.324} & \bluebox{31.324} \\  
\midrule\midrule
\multirow{40}{*}{Phi3.5 (3.8B)} & w/o RAG & 5.624 & -- & -- & -- & 51.518 & -- & -- & --\\
\cmidrule(lr){2-10}
 & Pre-Q & -- & \bluebox{9.786} & \bluebox{10.378} & \bluebox{33.709} & -- & \redbox{86.083} & \redbox{84.950} & \redbox{51.813} \\  
 & Aft-Q & -- & \bluebox{7.636} & \bluebox{9.270} & \bluebox{43.415} & -- & \redbox{88.486} & \redbox{85.947} & \bluebox{39.393} \\  
 & Aft-C & -- & \bluebox{7.682} & \bluebox{15.952} & \bluebox{42.476} & -- & \redbox{87.353} & \redbox{76.111} & \bluebox{44.334} \\  
\midrule
\multirow{40}{*}{Qwen2.5 (14B)} & w/o RAG & 12.125 & -- & -- & -- & 49.151 & -- & -- & --\\
\cmidrule(lr){2-10}
 & Pre-Q & -- & \redbox{8.646} & \redbox{8.740} & \redbox{11.892} & -- & \redbox{89.483} & \redbox{88.441} & \bluebox{47.280} \\  
 & Aft-Q & -- & \redbox{7.013} & \redbox{7.257} & \bluebox{17.592} & -- & \redbox{89.121} & \redbox{87.534} & \bluebox{40.798} \\  
 & Aft-C & -- & \redbox{7.746} & \redbox{9.778} & \redbox{8.531} & -- & \redbox{79.329} & \redbox{75.884} & \bluebox{45.014} \\  
\midrule
\multirow{40}{*}{Qwen2.5 (72B)} & w/o RAG & 4.030 & -- & -- & -- & 60.483 & -- & -- & --\\
\cmidrule(lr){2-10}
 & Pre-Q & -- & \bluebox{9.393} & \bluebox{7.896} & \bluebox{20.412} & -- & \redbox{89.982} & \redbox{85.766} & \bluebox{45.739} \\  
 & Aft-Q & -- & \bluebox{8.782} & \bluebox{9.781} & \bluebox{18.652} & -- & \redbox{93.246} & \redbox{89.574} & \bluebox{44.696} \\  
 & Aft-C & -- & \bluebox{9.270} & \bluebox{5.990} & \bluebox{23.564} & -- & \redbox{88.622} & \redbox{79.284} & \bluebox{39.483} \\  
    \bottomrule
    \end{tabular}
    }
    \caption{Evaluation results with MedMCQA. \redbox{Red} highlights areas where performance improved compared to the non-RAG setting, while \bluebox{Blue} indicates areas where performance deteriorated.}
    \label{tab:medmcqa-ece-ace}
\end{table*}

%% file: tables/pubmedqa-ece-ace-results.tex
\begin{table*}[t!]
\centering
\small
    \resizebox{0.98\textwidth}{!}{
    \renewcommand{\arraystretch}{0.1}
    \begin{tabular}{cc cccc cccc}
    \toprule
    \multirow{2}{*}{\textbf{Model}} & \multirow{2}{*}{\textbf{Pattern}} & \multicolumn{4}{c}{\textbf{ACE $\downarrow$}} & \multicolumn{4}{c}{\textbf{Accuracy $\uparrow$}} \\
    \cmidrule(lr){3-6}    \cmidrule(lr){7-10}
    & & \textbf{None} & \textbf{Ans1} & \textbf{Ans1-Oth2} & \textbf{Oth3} & \textbf{None} & \textbf{Ans1} & \textbf{Ans1-Oth2} & \textbf{Oth3}  \\
    \midrule
\multirow{40}{*}{Llama2 (70B)} & w/o RAG & 12.107 & -- & -- & -- & 46.400 & -- & -- & --\\
\cmidrule(lr){2-10}
 & Pre-Q & -- & \bluebox{29.791} & \bluebox{30.422} & \bluebox{14.146} & -- & \redbox{82.200} & \redbox{79.800} & \redbox{56.500} \\  
 & Aft-Q & -- & \bluebox{30.380} & \bluebox{31.220} & \bluebox{13.234} & -- & \redbox{81.600} & \redbox{74.100} & \redbox{53.000} \\  
 & Aft-C & -- & \bluebox{13.494} & \bluebox{13.430} & \redbox{11.322} & -- & \redbox{57.200} & \redbox{53.100} & \redbox{50.700} \\  
\midrule
\multirow{40}{*}{Llama3.1 (70B)} & w/o RAG & 6.091 & -- & -- & -- & 58.600 & -- & -- & --\\
\cmidrule(lr){2-10}
 & Pre-Q & -- & \bluebox{11.329} & \bluebox{11.513} & \bluebox{11.521} & -- & \bluebox{55.200} & \bluebox{55.200} & \bluebox{55.200} \\  
 & Aft-Q & -- & \bluebox{11.343} & \bluebox{11.532} & \bluebox{11.539} & -- & \bluebox{55.200} & \bluebox{55.200} & \bluebox{55.200} \\  
 & Aft-C & -- & \bluebox{11.370} & \bluebox{11.534} & \bluebox{11.543} & -- & \bluebox{55.200} & \bluebox{55.200} & \bluebox{55.200} \\  
\midrule
\multirow{40}{*}{Llama3.1 (8B)} & w/o RAG & 24.939 & -- & -- & -- & 11.000 & -- & -- & --\\
\cmidrule(lr){2-10}
 & Pre-Q & -- & \redbox{23.683} & \redbox{23.368} & \redbox{23.975} & -- & \redbox{12.200} & \redbox{12.600} & \redbox{12.000} \\  
 & Aft-Q & -- & \redbox{23.085} & \redbox{23.576} & \redbox{23.988} & -- & \redbox{12.800} & \redbox{12.400} & \redbox{12.000} \\  
 & Aft-C & -- & \redbox{23.930} & \redbox{23.854} & \redbox{24.370} & -- & \redbox{11.900} & \redbox{12.100} & \redbox{11.600} \\  
\midrule
\multirow{40}{*}{Meditron (70B)} & w/o RAG & 18.115 & -- & -- & -- & 34.800 & -- & -- & --\\
\cmidrule(lr){2-10}
 & Pre-Q & -- & \redbox{11.540} & \bluebox{18.483} & \redbox{8.365} & -- & \redbox{57.300} & \redbox{69.800} & \redbox{57.200} \\  
 & Aft-Q & -- & \redbox{9.159} & \redbox{6.645} & \redbox{6.270} & -- & \redbox{56.700} & \redbox{55.600} & \redbox{54.800} \\  
 & Aft-C & -- & \redbox{4.171} & \redbox{5.050} & \redbox{7.915} & -- & \redbox{54.700} & \redbox{54.800} & \redbox{55.100} \\  
\midrule
\multirow{40}{*}{PMC-Llama (13B)} & w/o RAG & 17.261 & -- & -- & -- & 22.800 & -- & -- & --\\
\cmidrule(lr){2-10}
 & Pre-Q & -- & \redbox{10.462} & \redbox{4.650} & \redbox{3.387} & -- & \redbox{28.800} & \redbox{37.900} & \redbox{36.600} \\  
 & Aft-Q & -- & \redbox{10.322} & \redbox{4.000} & \redbox{3.985} & -- & \redbox{28.900} & \redbox{39.200} & \redbox{40.000} \\  
 & Aft-C & -- & \redbox{4.169} & \redbox{5.421} & \redbox{7.250} & -- & \redbox{41.200} & \redbox{44.500} & \redbox{46.100} \\  
\midrule
\multirow{40}{*}{Gemma2 (2B)} & w/o RAG & 6.387 & -- & -- & -- & 55.200 & -- & -- & --\\
\cmidrule(lr){2-10}
 & Pre-Q & -- & \redbox{5.794} & \redbox{5.409} & \redbox{5.394} & -- & \redbox{55.300} & \redbox{55.200} & \redbox{55.200} \\  
 & Aft-Q & -- & \redbox{6.159} & \redbox{5.098} & \redbox{4.188} & -- & \redbox{55.200} & \redbox{55.200} & \redbox{55.200} \\  
 & Aft-C & -- & \bluebox{9.081} & \redbox{6.161} & \redbox{6.376} & -- & \redbox{55.200} & \redbox{55.200} & \redbox{55.200} \\  
\midrule\midrule
\multirow{40}{*}{Phi3.5 (3.8B)} & w/o RAG & 48.176 & -- & -- & -- & 33.400 & -- & -- & --\\
\cmidrule(lr){2-10}
 & Pre-Q & -- & \redbox{14.640} & \redbox{14.777} & \bluebox{57.831} & -- & \redbox{81.600} & \redbox{81.200} & \bluebox{21.900} \\  
 & Aft-Q & -- & \redbox{13.677} & \redbox{31.960} & \bluebox{52.083} & -- & \redbox{82.300} & \redbox{62.700} & \redbox{41.300} \\  
 & Aft-C & -- & \redbox{16.771} & \redbox{33.297} & \redbox{47.123} & -- & \redbox{73.700} & \redbox{52.300} & \redbox{33.800} \\  
\midrule
\multirow{40}{*}{Qwen2.5 (14B)} & w/o RAG & 15.874 & -- & -- & -- & 42.800 & -- & -- & --\\
\cmidrule(lr){2-10}
 & Pre-Q & -- & \redbox{4.746} & \redbox{4.816} & \bluebox{18.425} & -- & \redbox{83.400} & \redbox{83.200} & \bluebox{32.600} \\  
 & Aft-Q & -- & \redbox{3.460} & \redbox{5.013} & \bluebox{26.783} & -- & \redbox{82.800} & \redbox{76.100} & \bluebox{33.900} \\  
 & Aft-C & -- & \redbox{7.616} & \redbox{3.088} & \bluebox{23.229} & -- & \redbox{74.500} & \redbox{63.900} & \bluebox{32.100} \\  
\midrule
\multirow{40}{*}{Qwen2.5 (72B)} & w/o RAG & 7.205 & -- & -- & -- & 46.400 & -- & -- & --\\
\cmidrule(lr){2-10}
 & Pre-Q & -- & \bluebox{10.477} & \redbox{3.801} & \bluebox{25.283} & -- & \redbox{74.900} & \redbox{78.100} & \bluebox{33.000} \\  
 & Aft-Q & -- & \bluebox{8.024} & \bluebox{10.931} & \bluebox{17.828} & -- & \redbox{80.300} & \redbox{71.200} & \bluebox{34.300} \\  
 & Aft-C & -- & \bluebox{8.877} & \redbox{6.995} & \bluebox{13.543} & -- & \redbox{76.800} & \redbox{71.000} & \bluebox{42.500} \\  
    \bottomrule
    \end{tabular}
    }
    \caption{Evaluation results on PubMedQA. \redbox{Red} highlights areas where performance improved compared to the non-RAG setting, while \bluebox{Blue} indicates areas where performance deteriorated.}
    \label{tab:pubmedqa-ece-ace}
\end{table*}

%% file: tables/medmcqa-ece-results.tex
\begin{table}[ht!]
    \centering
    \renewcommand{\arraystretch}{0.1}
    \resizebox{!}{0.9\columnwidth}{
    \begin{tabular}{cc cccc}
    \toprule
    \multirow{2}{*}{\textbf{Model}} & \multirow{2}{*}{\textbf{Pattern}} & \multicolumn{4}{c}{\textbf{ECE $\downarrow$}} \\
    \cmidrule(lr){3-6} 
    & & \textbf{None} & \textbf{Ans1} & \textbf{Ans1-Oth2} & \textbf{Oth3} \\
\midrule
\multirow{35}{*}{Llama2 (70B)} & w/o RAG & 0.02 & -- & -- & -- \\
\cmidrule(lr){2-6}
& Pre-Q & -- & \bluebox{0.22} & \bluebox{0.25} & \bluebox{0.08} \\  
& Aft-Q & -- & \bluebox{0.24} & \bluebox{0.21} & \bluebox{0.12} \\  
& Aft-C & -- & \bluebox{0.20} & \bluebox{0.17} & \bluebox{0.10} \\  
\midrule
\multirow{35}{*}{Llama3.1 (70B)} & w/o RAG & 0.20 & -- & -- & -- \\
\cmidrule(lr){2-6}
& Pre-Q & -- & \redbox{0.14} & \redbox{0.14} & \redbox{0.14} \\  
& Aft-Q & -- & \redbox{0.14} & \redbox{0.14} & \redbox{0.14} \\  
& Aft-C & -- & \redbox{0.14} & \redbox{0.14} & \redbox{0.14} \\  
\midrule
\multirow{35}{*}{Llama3.1 (8B)} & w/o RAG & 0.03 & -- & -- & -- \\
\cmidrule(lr){2-6}
& Pre-Q & -- & \redbox{0.02} & \redbox{0.02} & \redbox{0.01} \\  
& Aft-Q & -- & \redbox{0.02} & \redbox{0.02} & \redbox{0.01} \\  
& Aft-C & -- & \redbox{0.02} & \redbox{0.02} & \redbox{0.01} \\  
\midrule
\multirow{35}{*}{Meditron (70B)} & w/o RAG & 0.07 & -- & -- & -- \\
\cmidrule(lr){2-6}
& Pre-Q & -- & \bluebox{0.18} & \bluebox{0.08} & \bluebox{0.09} \\  
& Aft-Q & -- & \bluebox{0.16} & \bluebox{0.09} & \bluebox{0.15} \\  
& Aft-C & -- & \bluebox{0.15} & \bluebox{0.07} & \bluebox{0.09} \\  
\midrule
\multirow{35}{*}{PMC-Llama (13B)} & w/o RAG & 0.16 & -- & -- & -- \\
\cmidrule(lr){2-6}
& Pre-Q & -- & \redbox{0.01} & \redbox{0.01} & \redbox{0.04} \\  
& Aft-Q & -- & \redbox{0.01} & \redbox{0.02} & \redbox{0.06} \\  
& Aft-C & -- & \redbox{0.02} & \redbox{0.05} & \redbox{0.05} \\  
\midrule
\multirow{35}{*}{Gemma2 (2B)} & w/o RAG & 0.20 & -- & -- & -- \\
\cmidrule(lr){2-6}
& Pre-Q & -- & \bluebox{0.24} & \bluebox{0.22} & \bluebox{0.21} \\  
& Aft-Q & -- & \bluebox{0.23} & \bluebox{0.21} & \bluebox{0.21} \\  
& Aft-C & -- & \bluebox{0.25} & \bluebox{0.23} & \bluebox{0.23} \\  
\midrule
\multirow{35}{*}{Phi-3.5 (3.8B)} & w/o RAG & 0.05 & -- & -- & -- \\
\cmidrule(lr){2-6}
& Pre-Q & -- & \bluebox{0.06} & \bluebox{0.07} & \bluebox{0.32} \\  
& Aft-Q & -- & \redbox{0.04} & \bluebox{0.06} & \bluebox{0.42} \\  
& Aft-C & -- & \redbox{0.04} & \bluebox{0.13} & \bluebox{0.41} \\  
\midrule
\multirow{35}{*}{Qwen2.5 (14B)} & w/o RAG & 0.12 & -- & -- & -- \\
\cmidrule(lr){2-6}
& Pre-Q & -- & \redbox{0.09} & \redbox{0.09} & \redbox{0.12} \\  
& Aft-Q & -- & \redbox{0.07} & \redbox{0.07} & \bluebox{0.18} \\  
& Aft-C & -- & \redbox{0.08} & \redbox{0.10} & \redbox{0.09} \\  
\midrule
\multirow{35}{*}{Qwen2.5 (72B)} & w/o RAG & 0.04 & -- & -- & -- \\
\cmidrule(lr){2-6}
& Pre-Q & -- & \bluebox{0.10} & \bluebox{0.08} & \bluebox{0.20} \\  
& Aft-Q & -- & \bluebox{0.09} & \bluebox{0.10} & \bluebox{0.19} \\  
& Aft-C & -- & \bluebox{0.10} & \bluebox{0.06} & \bluebox{0.24} \\  
\bottomrule
     \end{tabular}
     }
    \caption{The result of ECE using MedMCQA}
    \label{tab:medmcqa-ece-only}
\end{table}

%% file: tables/pubmedqa-ece-results.tex
\begin{table}[ht!]
    \centering
    \renewcommand{\arraystretch}{0.1}
    \resizebox{!}{0.9\columnwidth}{
    \begin{tabular}{cc cccc}
    \toprule
    \multirow{2}{*}{\textbf{Model}} & \multirow{2}{*}{\textbf{Pattern}} & \multicolumn{4}{c}{\textbf{ECE $\downarrow$}} \\
    \cmidrule(lr){3-6} 
    & & \textbf{None} & \textbf{Ans1} & \textbf{Ans1-Oth2} & \textbf{Oth3} \\
\midrule
\multirow{35}{*}{Llama2 (70B)} & w/o RAG & 0.12 & -- & -- & -- \\
\cmidrule(lr){2-6}
& Pre-Q & -- & \bluebox{0.30} & \bluebox{0.31} & \bluebox{0.14} \\  
& Aft-Q & -- & \bluebox{0.30} & \bluebox{0.31} & \bluebox{0.14} \\  
& Aft-C & -- & \bluebox{0.14} & \bluebox{0.14} & \bluebox{0.12} \\  
\midrule
\multirow{35}{*}{Llama3.1 (70B)} & w/o RAG & 0.02 & -- & -- & -- \\
\cmidrule(lr){2-6}
& Pre-Q & -- & \bluebox{0.10} & \bluebox{0.10} & \bluebox{0.10} \\  
& Aft-Q & -- & \bluebox{0.10} & \bluebox{0.10} & \bluebox{0.10} \\  
& Aft-C & -- & \bluebox{0.10} & \bluebox{0.10} & \bluebox{0.10} \\  
\midrule
\multirow{35}{*}{Llama3.1 (8B)} & w/o RAG & 0.24 & -- & -- & -- \\
\cmidrule(lr){2-6}
& Pre-Q & -- & \redbox{0.23} & \redbox{0.22} & \redbox{0.23} \\  
& Aft-Q & -- & \redbox{0.22} & \redbox{0.23} & \redbox{0.23} \\  
& Aft-C & -- & \redbox{0.23} & \redbox{0.23} & \redbox{0.23} \\  
\midrule
\multirow{35}{*}{Meditron (70B)} & w/o RAG & 0.18 & -- & -- & -- \\
\cmidrule(lr){2-6}
& Pre-Q & -- & \redbox{0.10} & \redbox{0.17} & \redbox{0.09} \\  
& Aft-Q & -- & \redbox{0.08} & \redbox{0.07} & \redbox{0.06} \\  
& Aft-C & -- & \redbox{0.04} & \redbox{0.04} & \redbox{0.07} \\  
\midrule
\multirow{35}{*}{PMC-Llama (13B)} & w/o RAG & 0.17 & -- & -- & -- \\
\cmidrule(lr){2-6}
& Pre-Q & -- & \redbox{0.09} & \redbox{0.01} & \redbox{0.01} \\  
& Aft-Q & -- & \redbox{0.09} & \redbox{0.01} & \redbox{0.02} \\  
& Aft-C & -- & \redbox{0.02} & \redbox{0.04} & \redbox{0.05} \\  
\midrule
\multirow{35}{*}{Gemma2 (2B)} & w/o RAG & 0.07 & -- & -- & -- \\
\cmidrule(lr){2-6}
& Pre-Q & -- & \redbox{0.04} & \redbox{0.02} & \redbox{0.02} \\  
& Aft-Q & -- & \redbox{0.02} & \redbox{0.01} & \redbox{0.01} \\  
& Aft-C & -- & \bluebox{0.09} & \redbox{0.05} & \redbox{0.07} \\  
\midrule
\multirow{35}{*}{Phi-3.5 (3.8B)} & w/o RAG & 0.48 & -- & -- & -- \\
\cmidrule(lr){2-6}
& Pre-Q & -- & \redbox{0.11} & \redbox{0.11} & \bluebox{0.58} \\  
& Aft-Q & -- & \redbox{0.10} & \redbox{0.28} & \bluebox{0.49} \\  
& Aft-C & -- & \redbox{0.15} & \redbox{0.32} & \redbox{0.46} \\  
\midrule
\multirow{35}{*}{Qwen2.5 (14B)} & w/o RAG & 0.16 & -- & -- & -- \\
\cmidrule(lr){2-6}
& Pre-Q & -- & \redbox{0.04} & \redbox{0.05} & \bluebox{0.18} \\  
& Aft-Q & -- & \redbox{0.03} & \redbox{0.05} & \bluebox{0.27} \\  
& Aft-C & -- & \redbox{0.07} & \redbox{0.03} & \bluebox{0.24} \\  
\midrule
\multirow{35}{*}{Qwen2.5 (72B)} & w/o RAG & 0.06 & -- & -- & -- \\
\cmidrule(lr){2-6}
& Pre-Q & -- & \bluebox{0.11} & \redbox{0.03} & \bluebox{0.25} \\  
& Aft-Q & -- & \bluebox{0.08} & \bluebox{0.11} & \bluebox{0.18} \\  
& Aft-C & -- & \bluebox{0.09} & \bluebox{0.07} & \bluebox{0.14} \\  
 \bottomrule
     \end{tabular}
     }
    \caption{The result of ECE using PubMedQA}
    \label{tab:pubmedqa-ece-only}
\end{table}